\documentclass[conference]{IEEEtran}
\usepackage{times}

\usepackage[numbers]{natbib}
\usepackage{multicol}
\usepackage[bookmarks=true]{hyperref}

\usepackage{graphicx}
\usepackage{amsfonts} 
\usepackage{amssymb}
\usepackage{amsmath}
\usepackage{array} 
\usepackage{booktabs}
\usepackage{multirow} 
\usepackage[table]{xcolor}
\usepackage{colortbl} 

\usepackage{marvosym}

\definecolor{paperbg}{HTML}{FBF6E8}
\newcommand{\tmeemethod}[1]{\cellcolor{paperbg}{#1}}
\newcommand{\tmeenum}[1]{\cellcolor{paperbg}\textbf{#1}}
\newcommand{\tmeenumno}[1]{\cellcolor{paperbg}{#1}}

\renewcommand{\footnoterule}{%
\kern -2pt
\hrule width 0.5\linewidth height 0.3pt
\kern 5pt
}

\begin{document}

\title{Reshaping Action Error Distributions for Reliable Vision-Language-Action Models}




%

\author{
Shuanghao Bai$^{1,2*}$ \quad
Dakai Wang$^{1*}$ \quad
Cheng Chi$^{2*}$ \quad
Wanqi Zhou$^{1}$ \quad
Jing Lyu$^{2,3,4}$ \quad
Xiaoguang Zhao$^{3}$ \\
Pengwei Wang$^{2}$ \quad
Zhongyuan Wang$^{2}$ \quad
Lei Xing$^{1}$  \quad
Shanghang Zhang$^{2,5\dagger}$ \quad
Badong Chen$^{1\dagger}$ \\[0.4em]


$^{1}$ Institute of Artificial Intelligence and Robotics, Xi'an Jiaotong University \\
$^{2}$ Beijing Academy of Artificial Intelligence \\
$^{3}$ Institute of Automation, University of Chinese Academy of Sciences \\
$^{4}$ School of Artificial Intelligence, University of Chinese Academy of Sciences \\
$^{5}$ Peking University

}


\maketitle

\begingroup
\renewcommand\thefootnote{}
\footnotetext{%
\textsuperscript{*} Equal contribution. \quad
\textsuperscript{$\dagger$} Corresponding authors.
}
\endgroup

\begin{abstract}
In robotic manipulation, vision–language–action (VLA) models have emerged as a promising paradigm for learning generalizable and scalable robot policies.
Most existing VLA frameworks rely on standard supervised objectives, typically cross-entropy for discrete actions and mean squared error (MSE) for continuous action regression, which impose strong pointwise constraints on individual predictions.
In this work, we focus on continuous-action VLA models and move beyond conventional MSE-based regression by reshaping action error distributions during training.
Drawing on information-theoretic principles, we introduce Minimum Error Entropy (MEE) into modern VLA architectures and propose a trajectory-level MEE objective, together with two weighted variants, combined with MSE for continuous-action VLA training.
We evaluate our approaches across standard, few-shot, and noisy settings on multiple representative VLA architectures, using simulation benchmarks such as LIBERO and SimplerEnv as well as real-world robotic manipulation tasks. Experimental results demonstrate consistent improvements in success rates and robustness across these settings. 
Under imbalanced data regimes, the gains persist within a well-characterized operating range, while incurring negligible additional training cost and no impact on inference efficiency.
We further provide theoretical analyses that explain why MEE-based supervision is effective and characterize its practical range.
Project Page: \href{https://cognition2actionlab.github.io/VLA-TMEE.github.io/}{VLA-TMEE Website}.
\end{abstract}

\IEEEpeerreviewmaketitle

\section{Introduction}

Information-theoretic principles have long served as a foundational driver of modern deep learning. In recent years, the rapid emergence of large-scale foundation models has further accelerated progress in artificial intelligence, leading to unprecedented growth across a wide range of domains.
In robotic manipulation, Vision-Language-Action (VLA) models have demonstrated strong potential for generalization and scalability by leveraging the powerful visual and linguistic understanding, reasoning, and generation capabilities inherited from large Vision-Language Models (VLMs)~\cite{zitkovich2023rt, kim2025openvla, bai2025towards}. Within this paradigm, classical information-theoretic learning objectives, such as cross entropy for classification and mean squared error (MSE) for regression, remain fundamental components. These objectives play central roles in discrete action modeling~\cite{ghosh2024octo, bu2025univla} and continuous action generation~\cite{reuss2024multimodal, fan2025long}, respectively.

\begin{figure*}[t]
\begin{center}
\centerline{\includegraphics[width=0.9\textwidth]{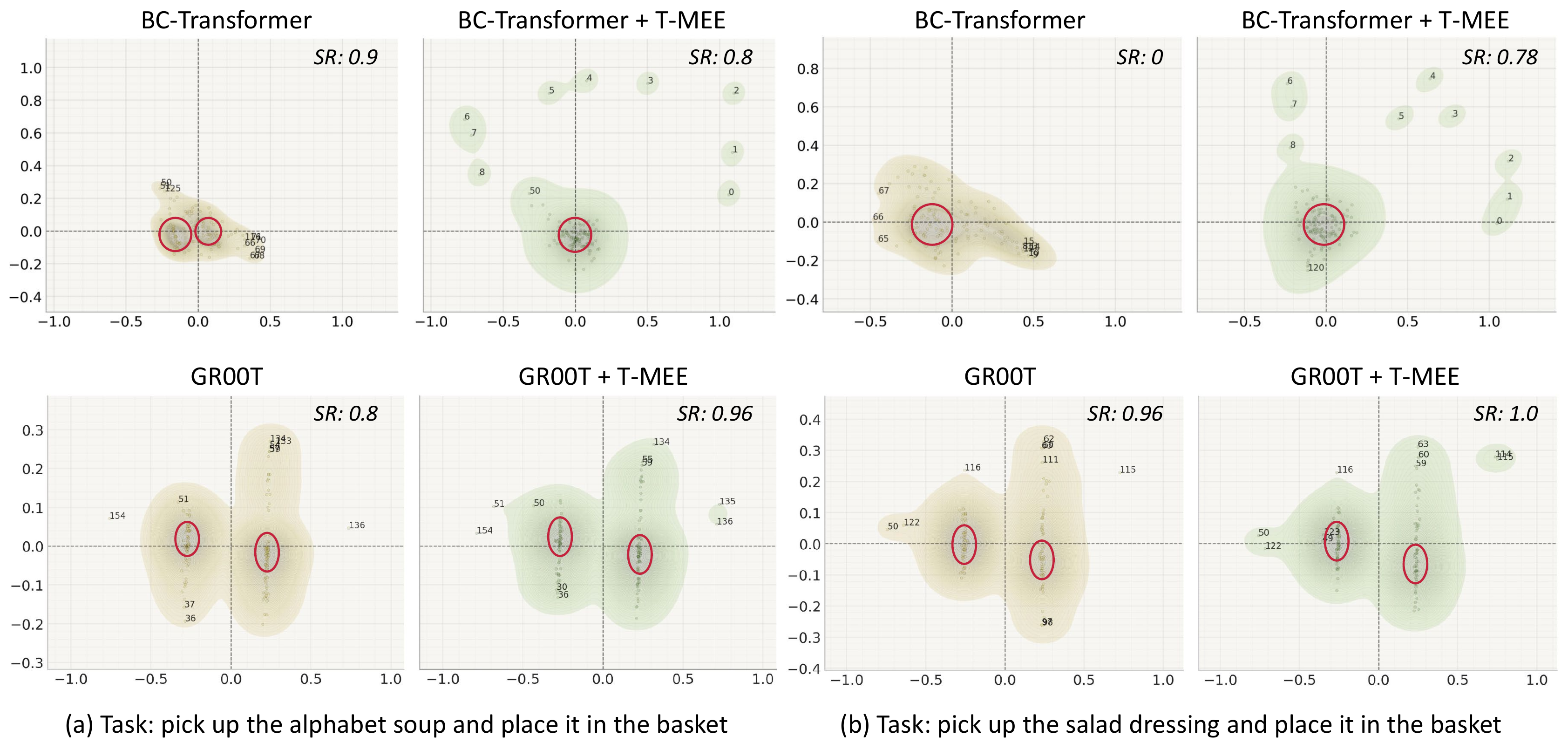}}
\vskip -0.1in
\caption{
\textbf{PCA visualization of action error distributions with and without trajectory-level MEE (T-MEE).}
Each point represents an action error at a specific time step along a trajectory. The top-10 most extreme outliers are highlighted with numeric labels, while red circles indicate compact action error clusters. 
Results are shown for BC-Transformer and GR00T trained with standard MSE-based behavior cloning and with the proposed T-MEE objective on LIBERO-Object. Per-task success rates (SR) for the two visualized tasks are annotated in the figure. For reference, the overall SR on the full 10-task LIBERO-Object suite improves from 57.4\% to 68.2\% for BC-Transformer and from 94.4\% to 97.8\% for GR00T. Across both architectures and tasks, incorporating T-MEE leads to more compact and coherent action error distributions in the projected space.
}
\label{fig: action_error}
\end{center}
\vskip -0.15in
\end{figure*}

As both discrete and continuous action modeling paradigms continue to evolve, we focus on continuous-action VLA models because they represent actions in a metric space with a well-defined notion of prediction error. This modeling choice naturally leads to pointwise regression objectives, most commonly mean squared error (MSE), which provide a simple and effective supervision signal for continuous control and are widely adopted in practice~\cite{li2024vision, kim2025fine}.
Despite their empirical success and stable optimization behavior, MSE-based objectives operate solely at the level of individual action predictions. From an analysis perspective, however, action prediction errors can be viewed not only as isolated deviations, but also as samples drawn from an underlying error distribution that evolves across time and action dimensions. Such distributions often exhibit structured patterns, including dispersion, skewness, or correlated variations, even when per-step regression losses are well optimized, as illustrated in Figure~\ref{fig: action_error}. This observation suggests that beyond per-step accuracy, the collective organization and geometry of action errors are not explicitly regulated by pointwise regression objectives.

From a distributional perspective, the behavior of a policy is governed not only by the magnitude of individual errors, but also by the uncertainty, concentration, and structure of the error distribution as a whole. Information-theoretic criteria provide a natural framework for characterizing such distribution-level properties, as they operate directly on probability distributions rather than individual samples. In particular, entropy-based objectives offer a principled way to quantify how concentrated or dispersed an error distribution is, without committing to strong parametric assumptions.
Among such objectives, the minimum error entropy (MEE) criterion~\cite{janzura1994minimum, chen2010mean} directly targets the entropy of the action error distribution, enabling interactions among errors within a batch or trajectory and providing a mechanism to aggregate errors beyond pointwise regression.

To operationalize entropy-based error aggregation in VLA models, we adapt the MEE criterion to the action prediction setting by reformulating it over trajectory-level action errors, resulting in Trajectory-level MEE (T-MEE). Building on this formulation, we further introduce two complementary weighted variants: Chunk-weighted T-MEE (Cw-TMEE) and Element-weighted T-MEE (Ew-TMEE). These variants provide finer-grained control over how action errors are aggregated and interact within the entropy objective.
We conduct a comprehensive empirical evaluation across multiple VLA architectures and model scales. Our experiments span near-balanced, few-shot, noisy, and imbalanced data regimes, and demonstrate consistent improvements across a broad range of settings. By systematically varying the degree of data imbalance, we further characterize the effective operating regime of entropy-based error aggregation and identify conditions under which it is most beneficial. Complementary analysis experiments provide insights into the resulting error distribution dynamics induced by the proposed objectives.

In addition to empirical validation, we provide a theoretical analysis of the optimization behavior induced by the MEE objective. Our analysis characterizes three key properties that help explain the observed empirical behavior. First, we show that MEE induces similarity-weighted interactions among action errors, resulting in a distribution-level optimization effect that regulates the relative geometry of the error space rather than penalizing individual error magnitudes. Second, we establish that the influence of outlying or highly corrupted errors is inherently bounded under MEE, offering a principled explanation for its robustness compared to conventional regression objectives. Finally, we analyze how MEE induces coupling or decoupling among multiple tasks through interactions in error space, and show that cross-task transfer and interference are jointly governed by task similarity, as reflected by the overlap of their error distributions, and by sample imbalance.

Our contributions are threefold. 
First, we revisit action errors from a distributional perspective and adapt MEE to the VLA setting, proposing three trajectory-level variants to capture structured action error distributions.
Second, we provide theoretical analyses that characterize the optimization behavior of the proposed objectives and establish conditions under which they are effective.
Finally, we conduct extensive empirical evaluations on two simulation benchmarks and real-robot manipulation tasks across multiple VLA architectures and model scales, spanning balanced, few-shot, noisy, and imbalanced training regimes, and complement these results with targeted analysis to validate effectiveness and provide insights into the behavior of the proposed methods.

\section{Related Work}

\noindent \textbf{Information-Theoretic Supervision for Continuous Regression.}
Mean squared error (MSE) is among the most widely used regression loss functions in machine learning due to its favorable optimization properties and stable behavior in practice~\cite{jadon2024comprehensive}. By penalizing squared deviations between predictions and targets, MSE enforces pointwise agreement but captures only a single aspect of prediction errors, namely their second-order moment, without explicitly modeling the global structure of the underlying error distribution. This limitation has motivated prior work on characterizing error distributions beyond pointwise loss measures~\cite{botchkarev2018performance}.
Information-theoretic objectives address this limitation by operating directly at the distribution level. In particular, Minimum Error Entropy (MEE)~\cite{janzura1994minimum, chen2010mean, chen2013kernel} minimizes the entropy of the empirical error distribution, encouraging errors to concentrate into compact and structured configurations without assuming a predefined parametric form. Compared to alternative information-theoretic criteria, such as KL divergence–based~\cite{zhang2025align} or mutual information–based~\cite{bai2025rethinking} objectives, MEE avoids the need to specify a target distribution or to perform explicit information estimation in high-dimensional continuous spaces.
In this work, we introduce MEE into the VLA setting and adapt it to continuous action regression. We further propose three trajectory-level variants and provide a comprehensive theoretical analysis to characterize their optimization behavior and effectiveness. A detailed review of related MEE literature is provided in Appendix~\ref{subsec: appendix_rw_mee}.

\noindent \textbf{Vision-Language-Action (VLA) Models.}
VLA models build upon advances in vision-language models, inheriting strong visual-language understanding, reasoning, and generation capabilities~\cite{zitkovich2023rt, kim2025openvla, zhao2025vlas}. These capabilities enable unified action modeling and have demonstrated favorable scaling behavior in robotic manipulation~\cite{ga2025gen}. Beyond large-scale foundation models, the VLA paradigm has also been extended to smaller-scale architectures that take visual-language inputs and directly produce action outputs, further broadening its applicability.
Existing VLA approaches broadly fall into two paradigms. The first formulates action generation as autoregressive token prediction, in which actions are discretized and generated in a token space~\cite{ghosh2024octo, wang2025vq}. The second focuses on continuous action generation, predicting continuous control signals via learnable action queries~\cite{bai2025rethinking, kim2025fine}, specialized action experts~\cite{li2024cogact, black2024pi0, bjorck2025gr00t, intelligence2025pi05}, or their combinations~\cite{wang2026vla}. This paradigm also includes dual-system VLA architectures, where a fast System~1 executes low-level actions at high frequency, while a slower System~2 provides contextual guidance over longer timescales~\cite{bu2024towards, cui2025openhelix}.
In this work, we focus on the continuous-action paradigm and study its modeling properties from the perspective of reshaping action error distributions. Across diverse architectures and model scales, we conduct extensive experiments to evaluate the effectiveness and applicability range of the proposed approach under varied training regimes and data conditions.

\section{Preliminaries}

\noindent \textbf{Problem Setting.}
We consider VLA learning under a behavior cloning (BC) paradigm for robotic manipulation. The environment is modeled as an instruction-conditioned sequential decision process, where a robot interacts with the environment under a natural language instruction.
At each timestep $t$, the robot receives a multimodal observation consisting of a visual observation $o^{t}$, an optional proprioceptive state $s^{t}$, and a language instruction $l$. We denote the overall input as $x^{t} = (o^{t}, s^{t}, l)$. The robot outputs a continuous action $a^{t} \in \mathcal{A}$, such as end-effector motion commands or arm-gripper controls.
In VLA, the policy $\pi_\theta$ is trained from an expert demonstration dataset $\mathcal{D}_e = \{(x^{t}, a^{t})\}$ by supervised learning. The objective is to learn a policy that mimics expert actions conditioned on the observed images and instructions. Formally, the BC objective is defined as
\begin{equation}
\pi^* = \arg\min_{\pi_\theta} \; \mathbb{E}_{(x^{t}, a^{t})\sim \mathcal{D}_e}
\big[ \mathcal{L}(\pi_\theta(x^{t}), a^{t}) \big],
\end{equation}
In continuous-action VLA, the loss $\mathcal{L}$ denotes a regression objective and is typically chosen as the MSE, yielding
\begin{equation}
\mathcal{L}_{\mathrm{MSE}} =
\mathbb{E}_{(x^{t}, a^{t})\sim \mathcal{D}_e}
\big[ \|\pi_\theta(x^{t}) - a^{t}\|^2 \big].
\end{equation}
This pointwise regression objective enforces per-timestep agreement between predicted and expert actions and serves as the standard training objective for continuous-action VLA models. In the following sections, we revisit action errors from a distributional perspective and introduce information-theoretic supervision to reshape action error distributions.

\noindent \textbf{Definition of MEE.}
In a standard regression setting, an input $x$ is mapped to an output by a parametric model $f_\theta$.
Let $y \in \mathbb{R}^d$ denote the ground-truth target, $\hat y = f_\theta(x)$ the prediction, and $e = y - \hat y$ the prediction error.
The Minimum Error Entropy (MEE) principle learns model parameters by minimizing the entropy of the error distribution.
When instantiated with Shannon entropy~\cite{shannon1948mathematical}, the objective is
\begin{equation}
\min_{\theta} \; H(e)
= - \int p(e) \log p(e)\, de ,
\end{equation}
where $p(e)$ denotes the probability density of the error variable.
Minimizing $H(e)$ encourages prediction errors to concentrate into a low-uncertainty distribution, rather than merely reducing their expected magnitude.
However, direct optimization of Shannon entropy is generally intractable due to the need for explicit density estimation.
In practice, MEE is commonly instantiated using the quadratic Rényi entropy~\cite{renyi1961measures},
\begin{equation}
H_2(e) = - \log \int p^2(e)\, de ,
\end{equation}
which provides a tractable approximation to Shannon entropy.
Minimizing $H_2(e)$ is equivalent to maximizing the squared density integral and favors compact error distributions.

\section{Method}

\begin{figure*}[t]
\begin{center}
\centerline{\includegraphics[width=0.85\textwidth]{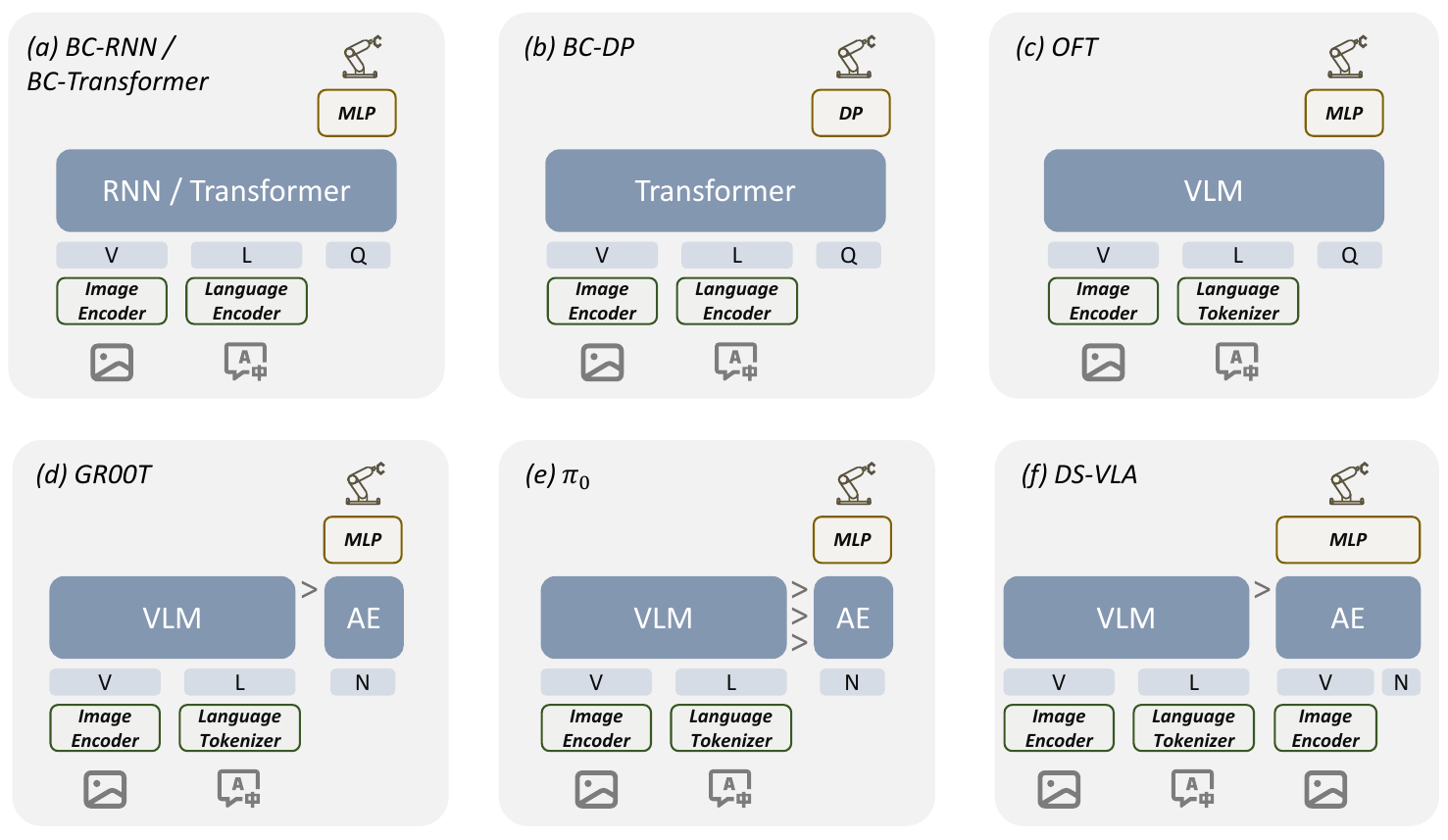}}
\vskip -0.1in
\caption{
\textbf{Architectural taxonomy of continuous-action VLA models evaluated in this work.}
We summarize representative small- and large-scale VLA architectures.
(a–b) Small-scale models regress actions from multimodal features using lightweight backbones: (a) BC-RNN / BC-Transformer with MLP policy heads, and (b) BC-DP with a diffusion-based action expert.
(c–f) Large-scale models build upon pretrained VLMs: (c) OFT introduces learnable action queries into autoregressive VLMs; (d) GR00T conditions an action expert (AE) on final-layer VLM features; (e) $\pi_0$ variants enable tighter VLM–action coupling via multi-layer conditioning or shared attention; and (f) DS-VLA adopts a dual-system design with a fast System~1 for action execution and a slower System~2 for contextual guidance.
Here, V denotes image tokens, L denotes language tokens, Q denotes query tokens, and N denotes noise inputs.
}
\label{fig: vla_type}
\end{center}
\vskip -0.15in
\end{figure*}

\subsection{Adapt MEE to VLA Models}
Classical MEE formulations are typically defined over scalar, sample-level prediction errors and are primarily applied in i.i.d. regression or adaptive filtering settings~\cite{chen2013kernel, chen2019minimum}. In contrast, action prediction in VLA models is structured, involving high-dimensional actions generated over temporally correlated trajectories. To bridge this gap, we reformulate MEE at the trajectory level by treating action prediction errors across time and batch as samples from a shared error distribution. This trajectory-level MEE (T-MEE) enables entropy-based supervision to operate on the collective geometry of action errors induced by VLA policies. Building on this formulation, we further introduce Chunk-weighted and Element-weighted T-MEE variants, which provide finer-grained control over error shaping across temporal segments and action dimensions.

In VLA models, let $\hat{\mathbf{a}}_{b,k}^{\,t} \in \mathbb{R}^D$ and $\mathbf{a}_{b,k}^{\,t}$ denote the predicted and ground-truth actions at the $k$-th step of an action chunk generated at time $t$ for trajectory $b$, where $k \in \{0,\dots,K-1\}$ and $K$ is the chunk size. We define the action prediction error as $\mathbf{e}_{b,k}^{\,t} = \hat{\mathbf{a}}_{b,k}^{\,t} - \mathbf{a}_{b,k}^{\,t}$.
Rather than treating errors at individual timesteps independently, we aggregate action prediction errors across batch, time, and chunk dimensions and regard them as samples drawn from a shared error random variable. This formulation aligns naturally with action chunking in VLA models, where multiple future actions are generated jointly as temporally correlated chunks.
Formally, given a batch of $B$ trajectories with horizon $T$ and chunk size $K$, the resulting set of error samples is
\begin{equation}
\mathcal{E} =
\left\{
\mathbf{e}_{b,k}^{\,t}
\;\middle|\;
b=1,\dots,B,\;
t=1,\dots,T,\;
k=0,\dots,K-1
\right\}.
\end{equation}

We then apply the quadratic Rényi entropy to the aggregated action error distribution.
Specifically, all action error vectors $\mathbf{e}_{b,k}^{\,t}$ across the batch, temporal, and chunk dimensions are flattened into a set of
$N = B \times T \times K$ samples, denoted as $\{\mathbf{e}_i\}_{i=1}^{N}$.
This yields the following T-MEE empirical objective:
\begin{equation}\label{eq: t-mee}
\mathcal{L}_{\mathrm{T\text{-}MEE}} =
- \log \left(
\frac{1}{N^2}
\sum_{i=1}^{N}
\sum_{j=1}^{N}
\exp\left(
- \frac{\| \mathbf{e}_i - \mathbf{e}_j \|^2}{2\sigma^2}
\right)
\right),
\end{equation}
where $\sigma$ denotes the kernel bandwidth. 
Following standard practice in information-theoretic learning and prior MEE-based methods,
$\sigma$ is chosen from the range $[0.5,2.0]$.

Building upon the T-MEE objective, we introduce a unified weighted
formulation that accounts for the varying reliability of action error samples.
Let $\{\mathbf{e}_i\}_{i=1}^{N}$ denote the aggregated action error vectors.
Each error sample is assigned a non-negative importance weight based on its magnitude:
\begin{equation}
w_i =
\frac{\exp\!\left(-\|\mathbf{e}_i\|^2 / 2\sigma_w^2\right)}
{\sum_{k=1}^{N} \exp\!\left(-\|\mathbf{e}_k\|^2 / 2\sigma_w^2\right)},
\end{equation}
where $\sigma_w$ denotes the kernel bandwidth, which is fixed to 0.5.
Using these weights, we define a weighted quadratic Rényi entropy estimator as
\begin{equation}\label{eq: mee-obj}
\mathcal{L}_{\mathrm{W\text{-}TMEE}}=
- \log
\sum_{i=1}^{N} \sum_{j=1}^{N}
\omega_{ij}
\exp\!\left(-\|\mathbf{e}_i - \mathbf{e}_j\|^2 / 2\sigma^2\right),
\end{equation}
where $\sigma$ denotes the kernel bandwidth and $\omega_{ij}$ specifies the weighting scheme.
We set $\omega_{ij} = \frac{1}{N^2} w_i$, yielding an asymmetric weighting that emphasizes reliable action chunks while aggregating errors across trajectories, which results in Chunk-weighted T-MEE (Cw-TMEE).
Alternatively, we define $\omega_{ij} = w_i w_j$, leading to a symmetric, element-weighted quadratic Rényi entropy estimator, referred to as Element-weighted T-MEE (Ew-TMEE).

Accurate trajectory imitation requires prediction errors to be not only structured but also centered near zero at each timestep. Pointwise regression supervision, such as MSE, provides this anchoring effect by directly penalizing the magnitude of individual action deviations and driving the error distribution toward the origin. 
Accordingly, we combine the distribution-level T-MEE objective with the standard MSE loss, yielding the final training objective:
\begin{equation}\label{eq: total_loss}
\mathcal{L}_{\mathrm{total}} = \mathcal{L}_{\mathrm{MSE}} + \alpha \, \mathcal{L}_{\mathrm{T\text{-}MEE}},
\end{equation}
where $\alpha$ balances the anchoring effect of pointwise accuracy and the distribution-level shaping.

\subsection{Model Architecture}
All evaluated models are summarized in Figure~\ref{fig: vla_type}. 
We consider both small-scale and large-scale VLA models.
For small-scale models, BC-RNN and BC-Transformer~\cite{liu2023libero} regress actions from multimodal features using lightweight sequence models with a shared perception backbone. Both adopt ResNet-18~\cite{he2016deep} as the image encoder and a BERT-based language encoder~\cite{devlin2019bert}. BC-RNN employs a 2-layer RNN backbone, while BC-Transformer uses a lightweight 4-layer Transformer~\cite{vaswani2017attention}, resulting in approximately 20M and 10M parameters, respectively. BC-DP~\cite{chi2025diffusion, bai2025rethinking} follows a similar perception setup but replaces the policy head with a diffusion-based action expert implemented as a 12-layer Transformer, yielding an overall model size of approximately 100M parameters.
For large-scale VLA models, we evaluate four representative architectural paradigms. OpenVLA-OFT-style methods regress continuous actions via action queries within autoregressive VLMs~\cite{kim2025fine}. GR00T-style models condition an action expert on final-layer VLM features~\cite{bjorck2025gr00t}, while $\pi_0$ variants explore tighter VLM--action coupling through multi-layer conditioning~\cite{black2024pi0}. DS-VLA adopts a dual-system design, where a fast System~1 executes actions at high frequency and a slower System~2 provides long-horizon contextual guidance~\cite{bu2024towards, cui2025openhelix}.
To ensure a fair comparison, we standardize the vision–language backbone across all large-scale models by replacing it with Qwen3-VL~\cite{bai2511qwen3} (using the 2B variant unless otherwise specified) and adopting its native image encoder.
Except for OFT, all large-scale models employ a shared flow-matching action head implemented as a 16-layer Transformer. As a result, all large-scale models have comparable capacity, with approximately 2.3B parameters each, enabling systematic evaluation of our method across diverse architectural designs.

\subsection{Theoretical Analysis}

This section analyzes the optimization behavior induced by the proposed T-MEE objective.
We identify three fundamental properties that explain how T-MEE shapes error interactions, improves robustness to non-Gaussian noise, and induces structured coupling in multi-task learning scenarios.

\subsubsection{Similarity-Weighted Interaction Between Trajectory Errors}
T-MEE operates in error space by inducing structured interactions among action prediction errors through pairwise similarity.
Let $\{\mathbf{e}_i\}_{i=1}^{N}$ denote the set of trajectory-level action prediction errors.
We define the Gaussian kernel measuring error similarity as:
\begin{equation}
k_{ij}
=
\exp\!\left(-\frac{\|\mathbf{e}_i-\mathbf{e}_j\|^2}{2\sigma^2}\right),
\quad
Z
=
\sum_{i=1}^{N}\sum_{j=1}^{N} k_{ij},
\end{equation}
where $Z$ denotes the corresponding information potential.
The T-MEE objective can be written compactly as
$\mathcal{L}_{\mathrm{T\text{-}MEE}} = -\log(Z/N^2)$.

\noindent\textbf{\textit{Proposition 1 (Similarity-Weighted Error Interaction).}}
\textit{Under the T-MEE objective, the negative gradient of each error sample induces a similarity-weighted interaction of the form:}
\begin{equation}
-\nabla_{\mathbf{e}_i}\mathcal{L}_{\mathrm{T\text{-}MEE}}
=
\frac{2}{\sigma^2 Z}
\sum_{j=1}^{N} k_{ij}(\mathbf{e}_j - \mathbf{e}_i).
\end{equation}
This result shows that T-MEE does not penalize errors independently.
Instead, each error sample is attracted toward other errors in proportion to their similarity, leading to clustering behavior in error space.
As a consequence, optimization under T-MEE regulates the collective geometry and entropy of the error distribution rather than merely reducing individual error magnitudes.
The proof is provided in Appendix~\ref{subsec: appendix_proof_p1}.

\subsubsection{Robustness to Non-Gaussian Noise and Outliers}
Beyond shaping error geometry, T-MEE exhibits inherent robustness to non-Gaussian noise and outliers due to its distribution-level formulation.

\noindent\textbf{\textit{Proposition 2 (Robustness via Higher-Order Statistics).}}
\textit{By minimizing Rényi’s quadratic entropy, T-MEE implicitly optimizes higher-order statistics of the error distribution, making it less sensitive to non-Gaussian perturbations than quadratic regression losses.}
The proof is provided in Appendix~\ref{subsec: appendix_proof_p2}.

\noindent\textbf{\textit{Proposition 3 (Bounded Influence of Outliers).}}
\textit{For an outlying error sample $\mathbf{e}_o$ far from the bulk of the error distribution, the gradient contribution induced by $\mathbf{e}_o$ under T-MEE is exponentially bounded. In particular,}
\begin{equation}
\bigl\|\nabla_{\mathbf{e}_o}
\mathcal{L}_{\mathrm{T\text{-}MEE}}\bigr\|
=
\mathcal{O}\!\left(c\,e^{-c^2/2}\right),
\end{equation}
\textit{where $c$ denotes the normalized distance to other error samples.}
In contrast to MSE, whose gradients grow linearly with error magnitude, the kernel-based weighting in T-MEE adaptively suppresses the influence of extreme deviations. As a result, the optimization dynamics are governed by the consensus structure of the error distribution rather than by isolated outliers. A full analysis is provided in Appendix~\ref{subsec: appendix_proof_p3}.

\subsubsection{Interaction Structure in Multi-Task Settings}
We analyze multi-task interaction under the T-MEE objective in the error space, where optimization dynamics are governed by the geometry of action prediction errors rather than task labels. 

\noindent \textbf{\textit{Proposition 4 (Imbalance-Induced Task Coupling).}} \textit{Consider two tasks $A$ and $B$ trained jointly under T-MEE. If their error distributions exhibit non-negligible overlap in error space, then sufficiently large sample imbalance causes the optimization of the minority task $B$ to be dominated by cross-task interactions from the majority task $A$. 
Conversely, when cross-task overlap is negligible, tasks evolve independently regardless of their relative sample sizes.} 

Multi-task coupling under T-MEE is jointly governed by \emph{task similarity} in error space and \emph{sample imbalance}, and can be quantified by the coupling ratio
\begin{equation}
R_B \;\triangleq\; 2 \cdot \frac{N_A}{N_B} \cdot \frac{\bar{k}_{AB}}{\bar{k}_{BB}},
\end{equation}
which characterizes the relative strength of cross-task interactions acting on the minority task $B$. When $R_B \gg 1$, task $B$ is dominated by majority-task errors, whereas $R_B \ll 1$ indicates effective decoupling. In benchmarks such as LIBERO, tasks sharing similar visual contexts and action primitives often exhibit substantial error-space overlap, and under severe data imbalance this leads to degraded performance on underrepresented tasks, consistent with our empirical observations. A formal analysis is provided in Appendix~\ref{subsec: appendix_proof_p4}.

\begin{table*}[t]
\centering
\caption{Performance on LIBERO across different model scales. We evaluate 40 tasks across four suites, each with 50 trials. Results are reported as success rates. Best results are bolded.}
\vskip -0.05in
\label{tab: libero}
\begin{tabular}{
>{\centering\arraybackslash}m{1.5cm}
>{\raggedright\arraybackslash}m{2.5cm}
>{\centering\arraybackslash}m{1.4cm}
>{\centering\arraybackslash}m{1.4cm}
>{\centering\arraybackslash}m{1.4cm}
>{\centering\arraybackslash}m{1.4cm}
>{\centering\arraybackslash}m{1.4cm}
}
\toprule
Params & Method & Spatial & Goal & Object & Long & Avg \\
\midrule

\multirow{4}{*}{\centering $<$20M}
& BC-RNN~\cite{liu2023libero}
& 40.0 & 8.2 & 11.0 & 0.8 & 15.0 \\
& \tmeemethod{\quad + T-MEE}
& \tmeenum{50.4} & \tmeenum{36.8} & \tmeenum{21.0} & \tmeenum{4.8} & \tmeenum{28.3} \\
& BC-Transformer~\cite{liu2023libero}
& 69.6 & 68.4 & 57.4 & 14.8 & 52.6 \\
& \tmeemethod{\quad + T-MEE}
& \tmeenum{85.2} & \tmeenum{73.2} & \tmeenum{68.2} & \tmeenum{27.2} & \tmeenum{63.5} \\
\midrule

\multirow{2}{*}{\centering $\sim$100M}
& BC-DP~\cite{chi2025diffusion}
& 65.6 & 79.2 & 91.2 & \textbf{80.2} & 79.1 \\
& \tmeemethod{\quad + T-MEE}
& \tmeenum{72.6} & \tmeenum{82.4} & \tmeenum{92.6} & \tmeenumno{80.0} & \tmeenum{81.9} \\
\midrule

\multirow{8}{*}{\centering $>$2B}
& GR00T~\cite{bjorck2025gr00t}
& 98.4 & 95.4 & 98.8 & 92.8 & 96.4 \\
& \tmeemethod{\quad + T-MEE}
& \tmeenum{98.8} & \tmeenum{96.2} & \tmeenum{99.4} & \tmeenum{93.4} & \tmeenum{97.0} \\
& OFT~\cite{kim2025fine}
& 98.8 & 93.6 & 98.4 & 91.2 & 95.5 \\
& \tmeemethod{\quad + T-MEE}
& \tmeenum{99.0} & \tmeenum{96.8} & \tmeenum{99.2} & \tmeenum{92.8} & \tmeenum{97.0} \\
& $\pi_0$~\cite{black2024pi0}
& 99.4 & 96.4 & 98.6 & 92.8 & 96.8 \\
& \tmeemethod{\quad + T-MEE}
& \tmeenum{99.8} & \tmeenum{98.2} & \tmeenum{100.0} & \tmeenum{95.6} & \tmeenum{98.4} \\
& DS-VLA~\cite{starvla2025}
& 98.2 & 96.4 & 99.2 & 89.2 & 95.8 \\
& \tmeemethod{\quad + T-MEE}
& \tmeenum{98.6} & \tmeenum{97.6} & \tmeenum{99.8} & \tmeenum{96.8} & \tmeenum{98.2} \\

\bottomrule
\end{tabular}
\end{table*}
\begin{table*}[t]
\centering
\caption{Performance of different VLA architectures across 2B and 4B base VLM scales on the SimplerEnv-WidowX benchmark.
Results are reported as success rates. Best results are bolded.}
\label{tab: simpler}
\vspace{-0.8em}

\begin{minipage}[t]{0.495\textwidth}
\centering
\footnotesize (a) 2B Base VLM

\vspace{0.35em}
\resizebox{0.99\linewidth}{!}{
\begin{tabular}{
>{\raggedright\arraybackslash}m{1.6cm}
>{\centering\arraybackslash}m{1cm}
>{\centering\arraybackslash}m{1cm}
>{\centering\arraybackslash}m{1cm}
>{\centering\arraybackslash}m{1.2cm}
>{\centering\arraybackslash}m{0.8cm}
}
\toprule
Method & Put Spoon & Put Carrot & Stack Cube & Put Eggplant & Avg \\
\midrule
GR00T~\cite{bjorck2025gr00t} & 87.5 & 54.2 & 12.5 & 45.8 & 50.0 \\
\rowcolor{paperbg}
\quad + T-MEE & 66.7 & 50.0 & 16.7 & 83.3 & \textbf{54.2} \\
\midrule
OFT~\cite{kim2025fine} & 37.5 & 25.0 & 0.0 & 91.7 & 38.5 \\
\rowcolor{paperbg}
\quad + T-MEE & 29.2 & 8.3 & 25.0 & 95.8 & \textbf{39.6} \\
\midrule
$\pi_0$~\cite{black2024pi0} & 37.5 & 25.0 & 4.2 & 70.8 & 43.8 \\
\rowcolor{paperbg}
\quad + T-MEE & 62.5 & 41.7 & 16.7 & 62.5 & \textbf{45.8} \\
\midrule
DS-VLA~\cite{starvla2025} & 58.3 & 54.2 & 12.5 & 70.8 & 49.0 \\
\rowcolor{paperbg}
\quad + T-MEE & 62.5 & 58.3 & 25.0 & 79.2 & \textbf{56.3} \\
\bottomrule
\end{tabular}
}
\end{minipage}
\hfill
\begin{minipage}[t]{0.495\textwidth}
\centering
\footnotesize (b) 4B Base VLM

\vspace{0.35em}
\resizebox{0.99\linewidth}{!}{
\begin{tabular}{
>{\raggedright\arraybackslash}m{1.6cm}
>{\centering\arraybackslash}m{1cm}
>{\centering\arraybackslash}m{1cm}
>{\centering\arraybackslash}m{1cm}
>{\centering\arraybackslash}m{1.2cm}
>{\centering\arraybackslash}m{0.8cm}
}
\toprule
Method & Put Spoon & Put Carrot & Stack Cube & Put Eggplant & Avg \\
\midrule
GR00T~\cite{bjorck2025gr00t} & 83.3 & 33.3 & 12.5 & 83.3 & 53.1 \\
\rowcolor{paperbg}
\quad + T-MEE & 75.0 & 45.8 & 16.7 & 95.8 & \textbf{58.3} \\
\midrule
OFT~\cite{kim2025fine} & 20.8 & 33.3 & 12.5 & 95.8 & 40.6 \\
\rowcolor{paperbg}
\quad + T-MEE & 45.8 & 33.3 & 4.2 & 83.3 & \textbf{41.7} \\
\midrule
$\pi_0$~\cite{black2024pi0} & 70.8 & 33.3 & 16.7 & 91.7 & 53.1 \\
\rowcolor{paperbg}
\quad + T-MEE & 79.2 & 41.7 & 33.3 & 91.7 & \textbf{61.5} \\
\midrule
DS-VLA~\cite{starvla2025} & 75.0 & 29.2 & 16.7 & 83.3 & 51.0 \\
\rowcolor{paperbg}
\quad + T-MEE & 75.0 & 45.8 & 12.5 & 91.7 & \textbf{56.3} \\
\bottomrule
\end{tabular}
}
\end{minipage}
\vskip -0.15in
\end{table*}

\section{Experiments}

\begin{figure*}[ht]
\begin{center}
\centerline{\includegraphics[width=1.0\textwidth]{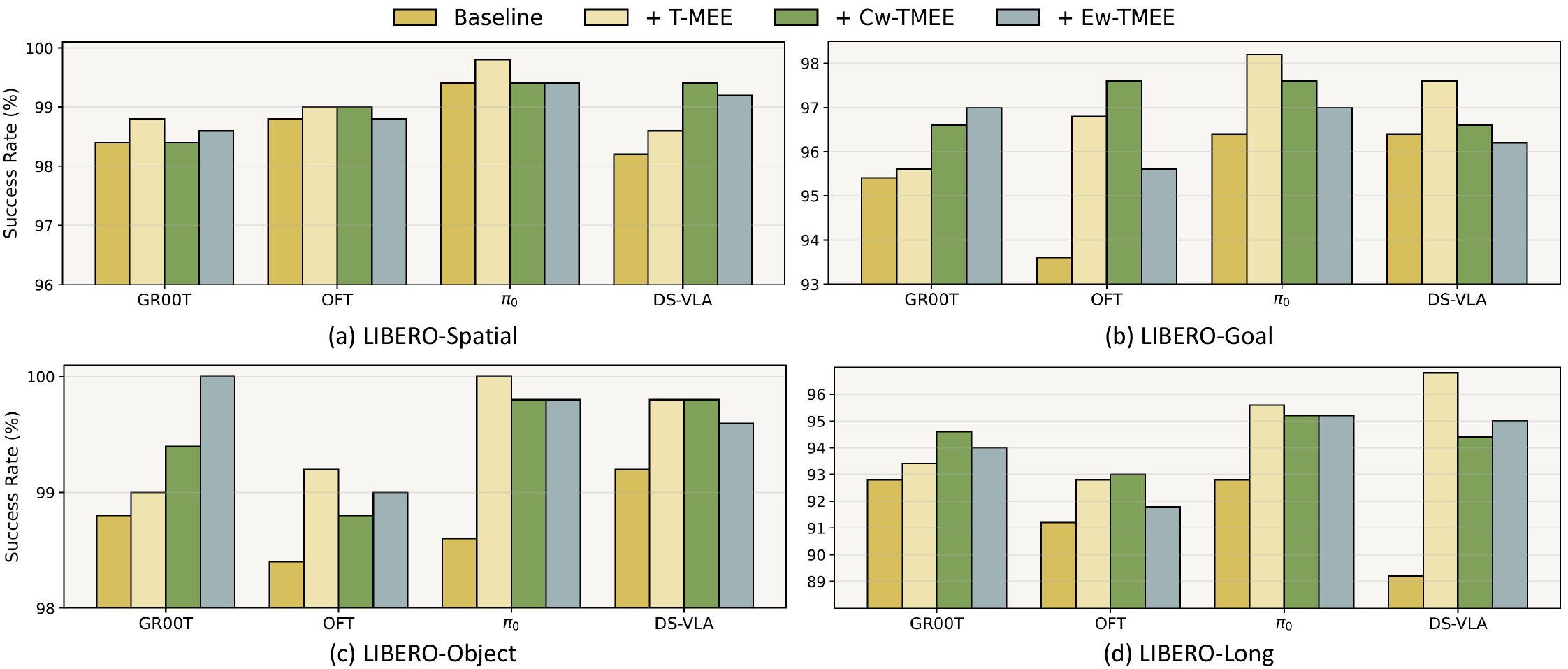}}
\vskip -0.1in
\caption{
\textbf{Performance comparison of MEE-based variants on LIBERO.}
We compare the baseline regression objective with three information-theoretic variants, including T-MEE, Chunk-weighted T-MEE (Cw-TMEE), and Element-weighted T-MEE (Ew-TMEE). Results are reported for representative continuous-action VLA architectures. All MEE-based objectives consistently improve performance over the baseline, while different variants exhibit complementary advantages across architectures and task suites, highlighting the flexibility of distribution-level error shaping.
}
\label{fig: mee_variants}
\end{center}
\vskip -0.2in
\end{figure*}
\begin{figure}[t]
\begin{center}
\centerline{\includegraphics[width=0.5\textwidth]{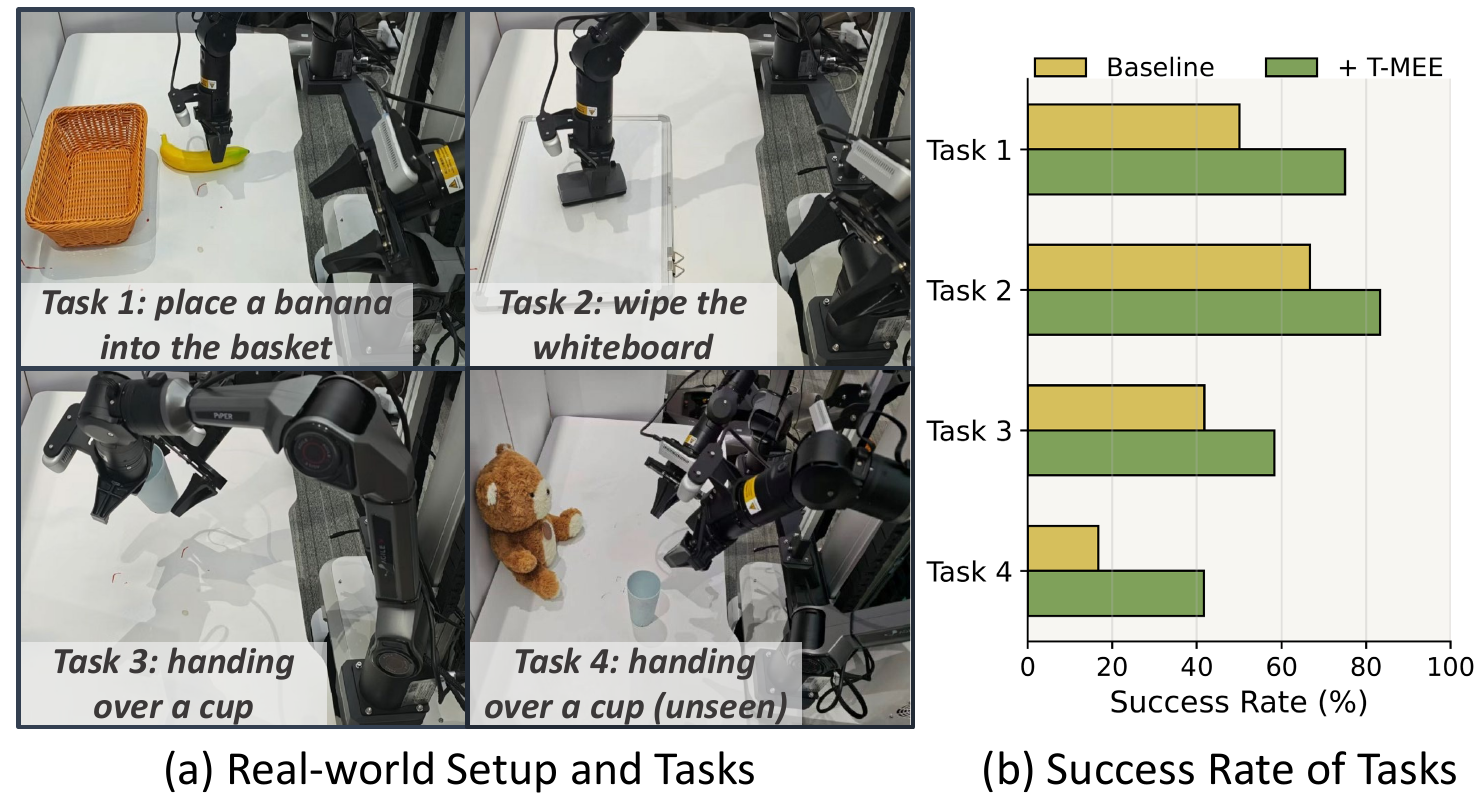}}
\vskip -0.05in
\caption{
Real-world evaluation.
(a) Real-world robotic setup and representative manipulation tasks.
(b) Task success rates comparing GR00T and GR00T + T-MEE, showing consistent performance gains from T-MEE across all tasks.
}
\label{fig: real_world}
\end{center}
\vskip -0.2in
\end{figure}

\subsection{Experiment Setup}

\noindent \textbf{Simulation Benchmark.} 
We evaluate our methods on two established benchmarks for robotic manipulation, LIBERO~\cite{liu2023libero} and SimplerEnv~\cite{li2025evaluating}.
We select four task suites from LIBERO, including Spatial, Goal, Object, and Long, each of which contains 10 single-arm manipulation tasks. For each suite, we report both suite-level success rates and the overall average, with 50 evaluation rollouts conducted per task.
SimplerEnv is a real-to-sim benchmark designed to measure the generalization and robustness of manipulation policies trained on real-world demonstrations. We evaluate on four WidowX manipulation tasks and report per-task success rates along with the overall average, based on 24 rollouts per task.

\noindent \textbf{Real-world Setup.} 
As shown in Figure~\ref{fig: real_world}, our real-world setup uses an Agilex Cobot Magic wheeled platform equipped with three RGB-D cameras. We consider three categories of manipulation tasks: placing a banana into the basket, wiping the whiteboard, and handing over a cup, along with one out-of-distribution task that involves handing over a cup in the presence of visual distractors. For data collection, we record 100 demonstration trajectories per task category at 30 Hz. During evaluation, each task is executed for 12 rollout trials.

\subsection{Main Results under Near-Balanced Data}

\noindent \textbf{Simulation Experiments.}
As shown in Table~\ref{tab: libero}, incorporating T-MEE consistently improves performance across all four suites. For small-scale models, including BC-RNN, BC-Transformer, and BC-DP, T-MEE yields substantial gains, improving the average success rate by up to +13.3\% for BC-RNN and +10.9\% for BC-Transformer. For large-scale VLA models, the improvements are more moderate but remain consistent, typically ranging from +0.5 to +2.4\%. As these models already achieve near-saturated performance, the marginal gains from further reshaping action error distributions are naturally smaller. Overall, these results demonstrate that T-MEE remains effective even under near-balanced data regimes.

\noindent \textbf{Real-world Robot Experiments.}
We select GR00T N1.5~\cite{bjorck2025gr00t} as the baseline for real-world evaluation. During real-robot experiments, we observe that incorporating T-MEE leads to noticeably \textbf{more stable and accurate} action execution, reflected in smoother trajectories and fewer corrective motions. As shown in Figure~\ref{fig: real_world}, T-MEE consistently improves the success rate of GR00T in real-world settings, indicating that the distribution-level regularization induced by T-MEE generalizes beyond simulation benchmarks and transfers effectively to physical robotic systems.

\noindent \textbf{Comparison of T-MEE Variants.}
As shown in Figure~\ref{fig: mee_variants}, all three T-MEE variants consistently outperform the baseline regression objective across multiple VLA architectures on LIBERO, demonstrating the effectiveness of distribution-level error supervision. Across architectures and evaluation suites, we find that the standard T-MEE objective accounts for most of the performance gains. While the weighted variants can provide additional improvements in certain settings, these gains are not consistent across models and tasks. In contrast, T-MEE exhibits robust performance across all architectures, indicating that T-MEE minimization alone is sufficient to effectively reshape action error distributions in practice.

\subsection{More Analyses}

\begin{figure*}[ht]
\begin{center}
\centerline{\includegraphics[width=0.9\textwidth]{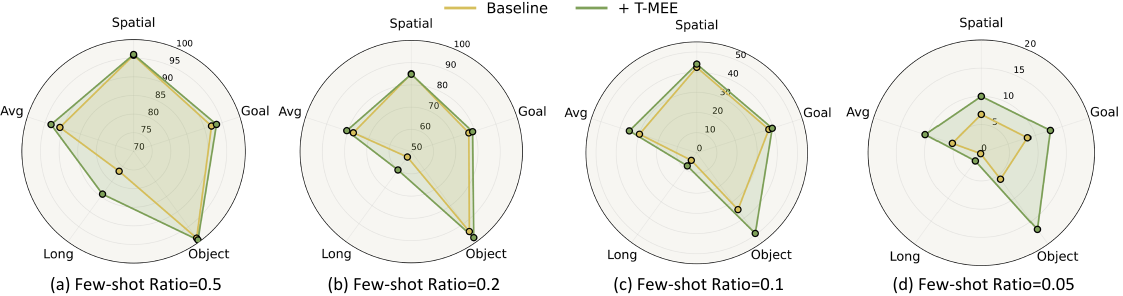}}
\vskip -0.05in
\caption{
Radar plots comparing GR00T and GR00T + T-MEE under different few-shot ratios. Each subplot reports success rates across the four LIBERO task suites and the overall average. As the amount of training data decreases, T-MEE consistently improves performance across task suites, with more pronounced gains in low-data regimes, indicating enhanced data efficiency and robustness to limited supervision.
}
\label{fig: few-shot_ridar}
\end{center}
\vskip -0.15in
\end{figure*}

\noindent \textbf{Few-Shot Learning.}
We evaluate T-MEE under few-shot supervision by varying the training data ratio on the LIBERO benchmarks. As shown in Figure~\ref{fig: few-shot_ridar}, T-MEE consistently improves success rates over the GR00T baseline across all task suites and data regimes, with more pronounced gains as the amount of training data decreases. These results indicate that T-MEE enhances data efficiency by providing distribution-level supervision, enabling more robust learning from limited demonstrations. Additional few-shot results across different models are provided in Appendix~\ref{subsec: appendix_few-shot}.

\begin{figure}[t]
\begin{center}
\centerline{\includegraphics[width=0.5\textwidth]{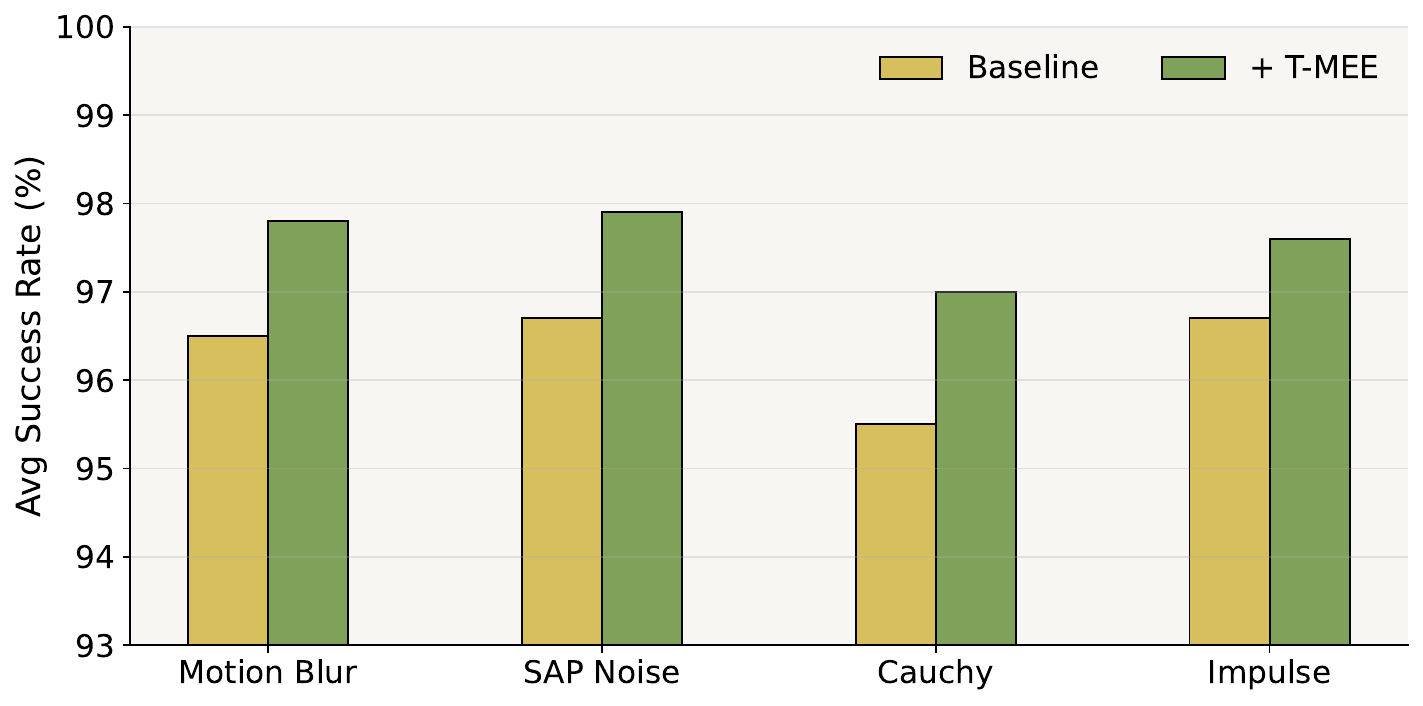}}
\vskip -0.05in
\caption{
Average success rates of GR00T with and without T-MEE under different noise corruptions on LIBERO.
}
\label{fig: noise_ft}
\end{center}
\vskip -0.25in
\end{figure}

\noindent \textbf{Robustness to Noises.}
The preceding theoretical analysis suggests that MEE is inherently robust to non-Gaussian noise and outliers. Accordingly, we consider two types of image corruptions, motion blur and salt-and-pepper noise, as well as two types of action noise, namely Cauchy noise and impulse noise. Detailed noise definitions and additional experiments are provided in Appendix~\ref{subsec: appendix_noise}. As shown in Figure~\ref{fig: noise_ft}, we fine-tune GR00T under each noise setting and evaluate performance on LIBERO. Across all noise types, incorporating T-MEE consistently improves the average success rate over the baseline. These results indicate that distribution-level supervision enhances robustness to corrupted observations and noisy demonstrations, providing a strong inductive bias for VLA training under noisy supervision.

\begin{figure}[t]
\begin{center}
\centerline{\includegraphics[width=0.48\textwidth]{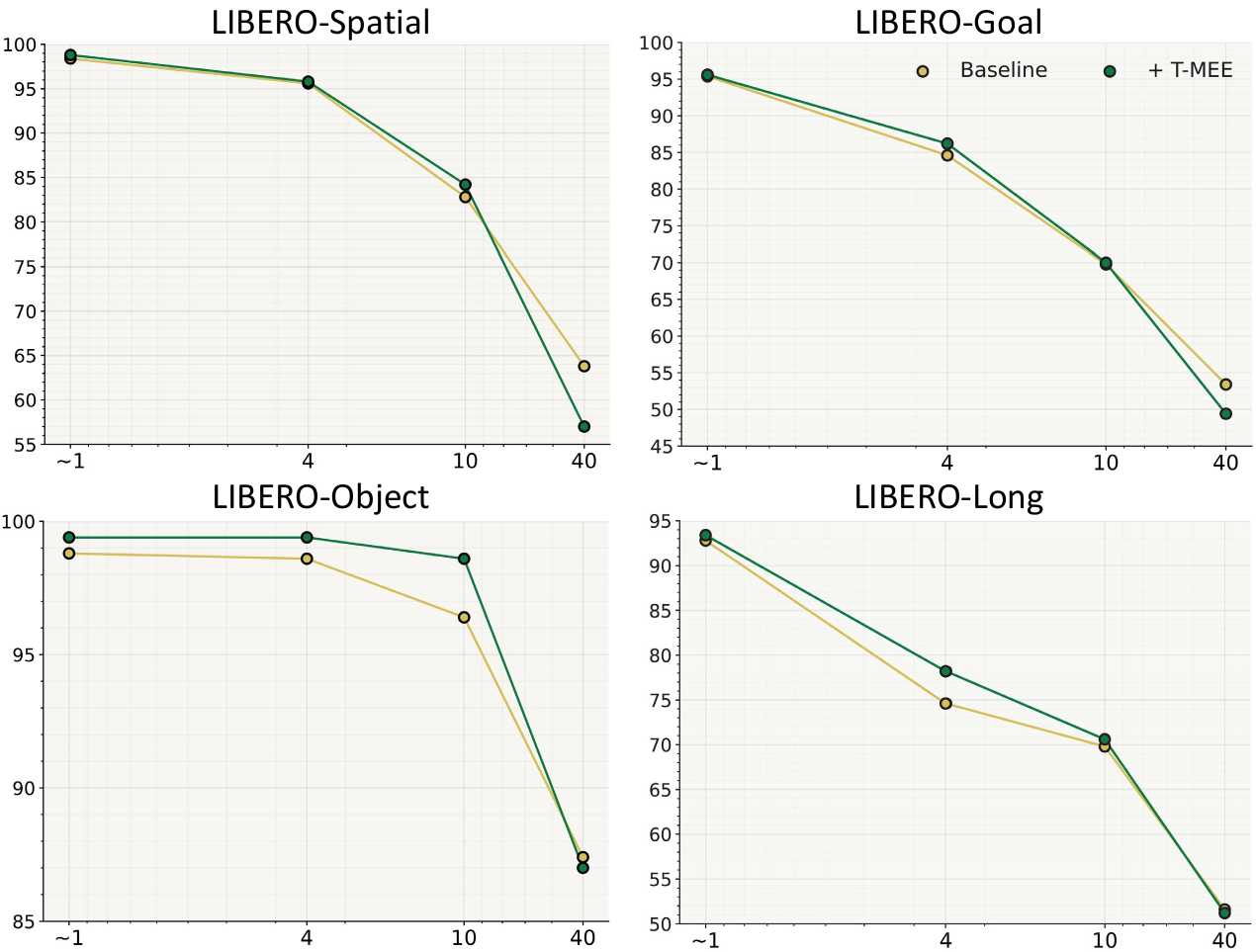}}
\vskip -0.05in
\caption{
Success rates of GR00T with and without T-MEE under different imbalance ratios on LIBERO.
}
\label{fig: imbalance_gr00t}
\end{center}
\vskip -0.25in
\end{figure}

\noindent \textbf{Imbalanced Data Regimes.}
Since each LIBERO suite consists of ten tasks with relatively balanced demonstrations, we construct imbalanced training regimes by assigning 40 demonstrations to five tasks as the majority group, while progressively reducing the remaining five tasks to 10, 4, and 1 demonstration, corresponding to imbalance ratios of 4, 10, and 40, respectively. As shown in Figure~\ref{fig: imbalance_gr00t}, we evaluate GR00T under varying degrees of task-level data imbalance.
Under mild to moderate imbalance, incorporating T-MEE consistently improves performance across all task suites. However, when the imbalance becomes extreme, corresponding to an imbalance ratio of 40, the benefits of T-MEE diminish and may no longer hold. This observation is consistent with our theoretical analysis, indicating that while T-MEE effectively mitigates moderate task imbalance by reshaping error distributions, its robustness has a practical limit when supervision for minority tasks becomes severely insufficient. These results characterize the effective operating range of T-MEE under imbalanced data regimes.
Notably, when evaluation focuses primarily on dominant tasks and does not explicitly probe long-tail tasks with very few demonstrations, performance gains can still be observed. As shown in Table~\ref{tab: simpler}, experiments on SimplerEnv further support this phenomenon. The Bridge dataset in SimplerEnv follows a highly long-tailed distribution, yet T-MEE continues to improve performance when evaluated on dominant tasks, suggesting that the benefits of T-MEE can manifest even when long-tail tasks are under-represented or not directly evaluated. Additional results under imbalanced data regimes are provided in Appendix~\ref{subsec: appendix_imbalance}.

\noindent \textbf{Error Analysis of One Trajectory.}
As shown in Figure~\ref{fig: action_error}, we observe a distinct behavior in small models. Action prediction errors at early timesteps exhibit pronounced outliers, whereas the aggregated error distribution over the full trajectory remains relatively compact and centered near zero. Despite these early outliers, the model attains a higher overall success rate. We attribute this phenomenon to limited model capacity. In the initial acceleration or initialization phase of the trajectory, smaller models may struggle to precisely predict action magnitudes, resulting in large instantaneous errors while still capturing correct directional trends.
In contrast, larger models do not exhibit this behavior. Their action errors remain tightly clustered throughout the trajectory, reflecting more consistent action execution and correspondingly higher task success.

\begin{figure}[t]
\begin{center}
\centerline{\includegraphics[width=0.5\textwidth]{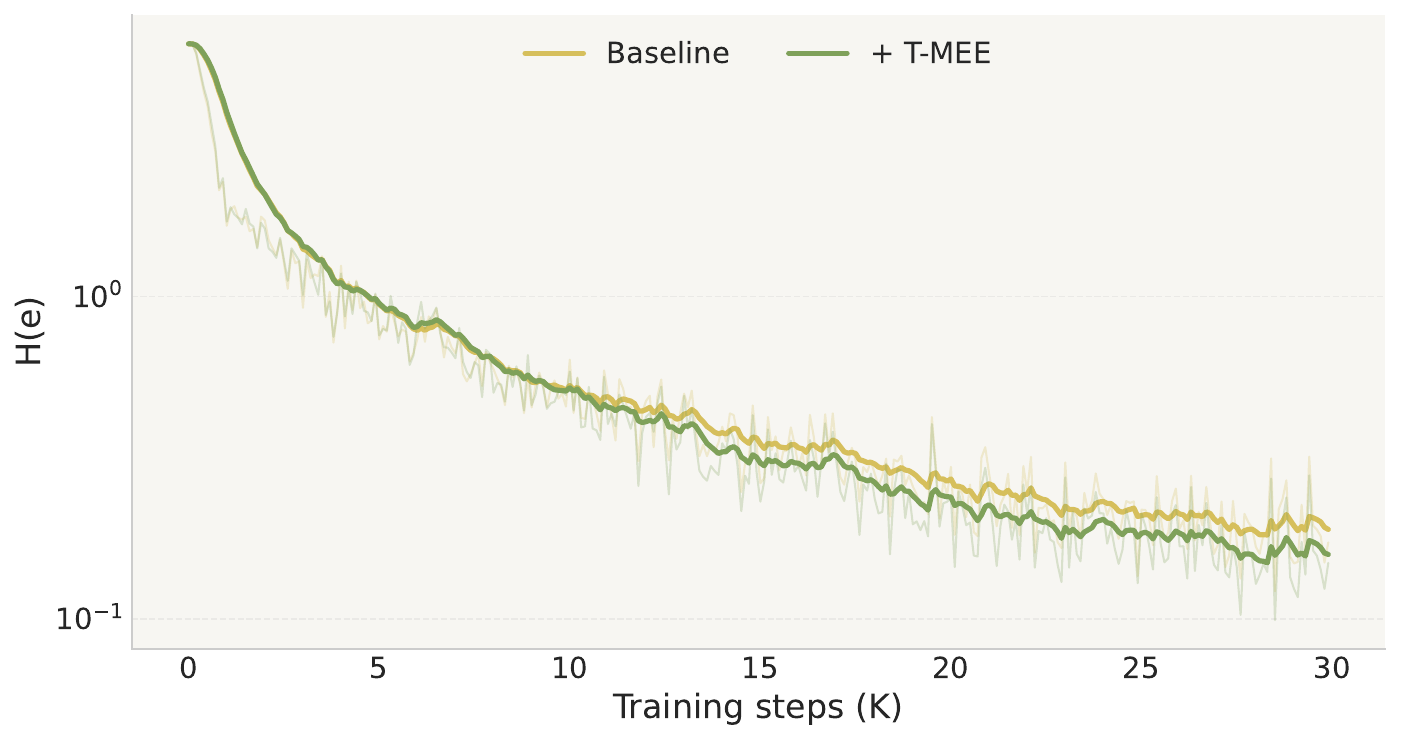}}
\vskip -0.05in
\caption{
Evolution of action error entropy during training.
}
\label{fig: error_entropy}
\end{center}
\vskip -0.35in
\end{figure}

\noindent \textbf{Error Entropy Visualizations.}
Figure~\ref{fig: error_entropy} visualizes the evolution of action error entropy during training.
Since T-MEE is introduced after 10k training steps, the two curves largely overlap in the early stage.
After T-MEE is activated, the entropy under T-MEE decreases more rapidly and converges to a lower level compared to the baseline.
This indicates that trajectory-level MEE effectively reshapes the error distribution once applied, leading to more compact and stable action errors.
Additional analyses are provided in Appendix~\ref{subsec: appendix_analyses}.

\section{Conclusion} 
\label{sec: conclusion}

We revisit action regression objectives for VLA models from an information-theoretic perspective and propose trajectory-level Minimum Error Entropy with two weighted variants. By reshaping action error distributions beyond pointwise regression, T-MEE promotes compact and structured errors without assuming a predefined parametric form. Experiments on simulation benchmarks and real-robot tasks demonstrate consistent improvements under standard, few-shot, noisy, and moderately imbalanced settings, with negligible training overhead. These results underscore the effectiveness of distribution-level supervision for enhancing robustness and data efficiency in continuous-action VLA models.

\bibliographystyle{plainnat}
\bibliography{references}

\newpage
\onecolumn

\appendix

\section{Appendix}

\subsection{Proof of Proposition 1}
\label{subsec: appendix_proof_p1}

\begin{proof}
Starting from the relation $\mathcal{L}_{\mathrm{T\text{-}MEE}} = -\log Z + \text{const}$, the gradient with respect to a specific error sample $\mathbf{e}_i$ is given by:
\begin{equation}
\nabla_{\mathbf{e}_i}\mathcal{L}_{\mathrm{T\text{-}MEE}} = - \frac{1}{Z}\nabla_{\mathbf{e}_i} Z.
\end{equation}
Considering the summation $Z = \sum_{a}\sum_{b} k_{ab}$, the gradient $\nabla_{\mathbf{e}_i} Z$ only involves terms where the index $a=i$ or $b=i$. Exploiting the symmetry of the Gaussian kernel ($k_{ij} = k_{ji}$), we obtain:
\begin{equation}
\nabla_{\mathbf{e}_i} Z = \sum_{j=1}^{N} \nabla_{\mathbf{e}_i} k_{ij} + \sum_{j=1}^{N} \nabla_{\mathbf{e}_i} k_{ji} = 2 \sum_{j=1}^{N} \left[ -\frac{k_{ij}}{\sigma^2}(\mathbf{e}_i - \mathbf{e}_j) \right].
\end{equation}
Substituting this result into the gradient of the loss and negating the sign yields the force $\mathbf{F}_i$:
\begin{equation}
\mathbf{F}_i = \frac{2}{\sigma^2 Z} \sum_{j=1}^{N} k_{ij}(\mathbf{e}_j - \mathbf{e}_i).
\end{equation}
This formulation reveals that each error $\mathbf{e}_i$ is ``pulled" toward other errors $\mathbf{e}_j$ with a strength proportional to their similarity $k_{ij}$, fostering a global clustering effect in the error space.
\end{proof}

\subsection{Proof of Proposition 2}
\label{subsec: appendix_proof_p2}

\begin{proof}
To analyze the statistical structure of the T-MEE objective, we consider
a Taylor expansion of the information potential
\begin{equation}
Z = \sum_{t=1}^{T}\sum_{s=1}^{T}
\exp\!\left(-\frac{\|e_t - e_s\|^2}{2\sigma^2}\right).
\end{equation}
with respect to the kernel bandwidth $\sigma$.
For sufficiently large $\sigma$ and any finite truncation order $n \ge 2$,
the exponential can be expanded as
\begin{equation}
\exp(-x)
= \sum_{k=0}^{n} \frac{(-1)^k}{k!}\, x^k
+ \mathcal{O}(x^{n+1}),
\qquad
x = \frac{\|e_t - e_s\|^2}{2\sigma^2}.
\end{equation}
Substituting $x$ and summing over $(t,s)$ gives
\begin{align}
Z
&= \sum_{t,s} \sum_{k=0}^{n}
    \frac{(-1)^k}{k!}
    \left(\frac{\|e_t - e_s\|^2}{2\sigma^2}\right)^k
  + \mathcal{O}(\sigma^{-2(n+1)}) \notag \\
&= T^2
 + \sum_{k=1}^{n} \frac{(-1)^k}{k!\,2^k\,\sigma^{2k}}
    \sum_{t,s} \|e_t - e_s\|^{2k}
 + \mathcal{O}(\sigma^{-2(n+1)}).
\label{eq:tmee_expansion_general}
\end{align}
In particular, the first two non-constant terms can be written explicitly as
\begin{align}
Z
&= T^2
   - \frac{1}{2\sigma^2}\sum_{t,s}\|e_t - e_s\|^2
   + \frac{1}{8\sigma^4}\sum_{t,s}\|e_t - e_s\|^4 \notag \\
&\quad
   + \sum_{k=3}^{n} \frac{(-1)^k}{k!\,2^k\,\sigma^{2k}}
     \sum_{t,s} \|e_t - e_s\|^{2k}
   + \mathcal{O}(\sigma^{-2(n+1)}).
\end{align}

The leading non-constant term recovers the MSE structure.
Using the identity
\begin{equation}
\sum_{t,s}\|e_t - e_s\|^2
= 2T \sum_{t=1}^{T}\|e_t - \bar e\|^2,
\qquad
\bar e = \frac{1}{T}\sum_{t} e_t,
\end{equation}
we see that minimizing the second-order term is equivalent to minimizing the empirical MSE loss.

The higher-order terms with $k \ge 2$ depend on
$\sum_{t,s}\|e_t - e_s\|^{2k}$ and are therefore sensitive to higher-order
moments (e.g., kurtosis) and heavy-tailed structure in the error distribution.
While a pure MSE objective only controls the second-order moment,
the T-MEE objective incorporates these higher-order contributions,
penalizing large pairwise deviations and encouraging trajectory errors to
concentrate in a compact region of the error manifold, thereby suppressing
stochastic outliers.
\end{proof}

\subsection{Proof of Proposition 3}
\label{subsec: appendix_proof_p3}

\textit{Consider an outlying error sample $\mathbf{e}_o$ far from the consensus, i.e., $\|\mathbf{e}_o-\mathbf{e}_j\| \ge c\sigma$ for all $j\neq o$ with $c\gg 1$. The gradient contribution induced by $\mathbf{e}_o$ is exponentially bounded, ensuring the stability of the optimization:}
\begin{equation}
\bigl\|\nabla_{\mathbf{e}_o} \mathcal{L}_{\mathrm{T\text{-}MEE}}\bigr\| = \Bigl\| \frac{2}{\sigma^2 Z} \sum_{j=1}^T k_{oj}(\mathbf{e}_o-\mathbf{e}_j) \Bigr\| = \mathcal{O}\!\left(c e^{-c^2/2}\right).
\end{equation}

Robustness is manifested through the \textit{adaptive importance weights} $k_{ij}$ within the gradient expression. In standard MSE, the gradient $\nabla \mathcal{L}_{\text{MSE}} \propto (\mathbf{e}_o - \mathbf{e}_j)$ grows linearly with the error magnitude, allowing a single outlier to dominate the parameter updates (the "leverage point" effect). In contrast, within the T-MEE framework, as the distance $\|\mathbf{e}_o - \mathbf{e}_j\|$ increases, the Gaussian kernel $k_{oj}$ decays exponentially. For $c \gg 1$, the product $k_{oj} \|\mathbf{e}_o - \mathbf{e}_j\|$ vanishes after reaching a finite maximum, effectively "rejecting" the outlier. Consequently, the optimization trajectory remains governed by the consensus of the error distribution rather than extreme values.

\subsection{Proof of Proposition 4}
\label{subsec: appendix_proof_p4}

In this section, we provide a formal analysis of the interaction structure
induced by the T-MEE objective in multi-task learning settings, and derive the
coupling criterion stated in Proposition~4.

\paragraph{Setup}
Consider two task groups $A$ and $B$, with $N_A$ and $N_B$ error samples,
respectively.
Let $\mathbf{e}_i$ denote an error sample, and let
\begin{equation}
k_{ij} = \exp\!\left(-\frac{\|\mathbf{e}_i - \mathbf{e}_j\|^2}{2\sigma^2}\right)
\end{equation}
denote the Gaussian kernel similarity between error samples $i$ and $j$.
Recall that the T-MEE objective is defined as
\begin{equation}
\mathcal{L}_{\mathrm{T\text{-}MEE}}
=
-\log\!\left(\frac{1}{N^2}\sum_{i,j} k_{ij}\right),
\end{equation}
where $N = N_A + N_B$.

\paragraph{Within- and cross-task interaction scaling}
We analyze the optimization dynamics acting on an error sample
$\mathbf{e}_i$ belonging to task $B$.
From the gradient expression of T-MEE, the force acting on $\mathbf{e}_i$
is proportional to a similarity-weighted sum over all other error samples.
This interaction naturally decomposes into two components:
(i) within-task interactions with other samples from task $B$, and
(ii) cross-task interactions with samples from task $A$.
To characterize their relative strength, we define the empirical average kernel similarities:
\begin{equation}\begin{aligned}
\bar{k}_{BB}
&=
\frac{1}{N_B^2}\sum_{i,j \in B} k_{ij}, \\
\bar{k}_{AB}
&=
\frac{1}{N_A N_B}\sum_{i \in A,\, j \in B} k_{ij}.
\end{aligned}\end{equation}

For an error sample $\mathbf{e}_i \in B$, the expected magnitude of within-task interactions scales as $\mathcal{O}\!\left(N_B\,\bar{k}_{BB}\right)$, while the expected magnitude of cross-task interactions scales as $\mathcal{O}\!\left(N_A\,\bar{k}_{AB}\right)$.
These scalings reflect the fact that each error sample interacts with all
other samples, weighted by their kernel similarity.

\paragraph{Coupling criterion}
Comparing the two interaction terms, the relative strength of cross-task
interactions acting on task $B$ is characterized by the ratio
\begin{equation}
R_B
=
2 \cdot \frac{N_A}{N_B} \cdot \frac{\bar{k}_{AB}}{\bar{k}_{BB}},
\label{eq:coupling-ratio-appendix}
\end{equation}
where the factor of $2$ accounts for symmetric cross-task contributions
arising from pairs $(i,j)\in A\times B$ and $(i,j)\in B\times A$ in the kernel
aggregation.

When $R_B \gg 1$, cross-task interactions from task $A$ dominate the
optimization dynamics of task $B$, causing the minority task to be strongly
coupled to the majority task.
In contrast, when $R_B \ll 1$, within-task interactions dominate and task $B$
evolves largely independently.
This establishes that multi-task coupling under T-MEE is jointly governed by
sample imbalance and the overlap of error distributions in error space, as
reflected by the kernel similarity ratio $\bar{k}_{AB}/\bar{k}_{BB}$.

\paragraph{Relation to kernel aggregation.}
For completeness, the kernel aggregation term admits the exact decomposition
\begin{equation}\label{eq: appendix_interaction_decomp}
\sum_{i,j} k_{ij}
=
N_A^2 \bar{k}_{AA}
+
2 N_A N_B \bar{k}_{AB}
+
N_B^2 \bar{k}_{BB},
\end{equation}
where $\bar{k}_{AA}$ denotes the average within-task similarity for task $A$.
This decomposition makes explicit how within-task and cross-task interactions
contribute to the overall information potential, but the coupling behavior of
individual tasks are governed by the relative scaling captured by
Equation~\eqref{eq: appendix_interaction_decomp}.

\subsection{Additional Related Work of Minimum Error Entropy}
\label{subsec: appendix_rw_mee}

Minimum Error Entropy (MEE) was originally introduced in the early 1990s as an estimation principle based on minimizing the entropy of the error distribution by Janzura et al.~\cite{janzura1994minimum}.
It was later developed into a learning criterion within the information-theoretic learning framework by Principe and collaborators~\cite{erdogmus2002error, principe2000information}, and subsequently further advanced 
in the context of robust learning and signal processing~\cite{chen2010mean}.
Prior work has established important theoretical properties of MEE, including its convergence behavior~\cite{chen2010mean, xie2020fixed} and robustness to non-Gaussian noise and outliers~\cite{chen2016insights, chen2019minimum, dang2020robust, he2023generalized, zhou2025information}.
By directly operating on the error distribution, MEE is more sensitive to higher-order statistical characteristics of the error signal, such as heavy tails and peakedness, and provides more stable gradient directions when noise deviates from Gaussian assumptions.
The computational complexity of MEE has also been studied extensively~\cite{chen2013kernel, chen2018quantized}.
Early work emphasized the quadratic complexity induced by kernel-based entropy estimation, particularly in online learning and adaptive filtering settings where model capacity was limited and the loss computation dominated the overall cost.
In contrast, in modern large-scale Vision--Language--Action models, the computational overhead of MEE-based objectives becomes negligible relative to the dominant cost of deep representation learning and action generation.
When implemented with GPU-parallelized tensor operations over mini-batches or trajectories, the additional cost introduced by MEE is effectively amortized by the forward and backward passes of the backbone network.
As a result, in the regime of large-scale VLA models, the primary challenge of applying MEE is no longer computational feasibility, but how to properly structure error interactions and integrate distribution-level supervision with modern action generation objectives.

\subsection{Implementation Details}
\label{subsec: appendix_implementation}

\subsubsection{Simulation Experiments}

\paragraph{Small-scale VLA models}
We evaluate T-MEE on three representative small-scale behavior cloning models, including BC-DP, BC-RNN, and BC-Transformer.
All models share the same visual encoder, language processing pipeline, data augmentation strategy, and evaluation protocol, while differing in temporal modeling and action head architectures.

\textbf{Shared Settings.}
All small-scale models share the same visual encoder, language processing pipeline, data augmentation strategy, and training protocol.
Specifically, we use a ResNet-based image encoder with the last four layers removed, trained from scratch without pretrained weights or frozen parameters.
Models are trained for 50 epochs with a batch size of 128 on a single A100 GPU, using the AdamW optimizer with a learning rate of $1\times10^{-4}$, $\beta=(0.9,0.999)$, and weight decay $1\times10^{-4}$.
Visual observations include agent-view and eye-in-hand RGB images at a resolution of $128\times128$, together with gripper and joint states.
Language instructions are encoded using a BERT-based tokenizer and fused into the visual stream via FiLM conditioning.
All models are evaluated over 50 rollout episodes per task.

\begin{table}[htbp]
\centering
\caption{Shared settings for small-scale VLA models.}
\label{tab: shared_small_model_settings}
\vskip -0.05in
\begin{tabular}{ll}
\toprule
\textbf{Category} & \textbf{Configuration} \\
\midrule
Image Encoder & ResNet, last 4 layers removed, trained from scratch \\
Image Resolution & $128 \times 128$ (agent-view, eye-in-hand) \\
Language Encoder & BERT tokenizer, max length 25 \\
Language Fusion & FiLM conditioning \\
Optimizer & AdamW ($1\times10^{-4}$, $\beta=(0.9,0.999)$) \\
Weight Decay & $1\times10^{-4}$ \\
Batch Size & 128 \\
Training Epochs & 50 \\
Hardware & Single NVIDIA A100 GPU \\
Data Augmentation & Brightness/contrast/saturation/hue (0.3), noise $\epsilon=0.1$ \\
Evaluation & 50 rollouts per task (LIBERO) \\
\bottomrule
\end{tabular}
\end{table}

\textbf{MEE-based Settings.}
During training, we adopt a two-phase optimization strategy.
In the first one-third of the total training steps, we optimize the model using only the MSE objective to stabilize the action error distribution.
In the remaining two-thirds of training, we activate the MEE-based objectives.
For T-MEE, the kernel bandwidth $\sigma$ is fixed to $0.5$, and the loss weight $\alpha$ is selected from $\{0.01, 0.1, 1.0\}$.
For the weighted variants, Cw-TMEE and Ew-TMEE, all hyperparameters follow the T-MEE setting, with the additional bandwidth parameter $\sigma_w$ also fixed to $0.5$.

\textbf{BC-DP.}
The BC-DP model adopts a diffusion-based action head with a DiT-B backbone.
The diffusion head operates on tokens of dimension 64, predicts 7-dimensional actions, and models a future action window of 9 steps (corresponding to an action chunk length of 10), with 8 repeated diffusion steps.
Temporal dependencies are modeled using a 4-layer transformer with 6 attention heads and a maximum sequence length of 10.
For T-MEE, the kernel bandwidth is fixed to $\sigma=0.5$, while the MEE loss weight is set to 0.01 for spatial and object tasks, and 0.1 for goal and long-horizon tasks.

\textbf{BC-RNN.}
The BC-RNN model employs a two-layer unidirectional RNN with a hidden size of 1024 to capture temporal dependencies.
Actions are predicted using a two-layer MLP head with a hidden size of 1024.
The kernel bandwidth is set to $\sigma=0.5$ for all task suites.
The T-MEE loss weight is set to 1.0 for spatial, goal, and long-horizon tasks, and 0.1 for object-centric tasks.

\textbf{BC-Transformer.}
The BC-Transformer model uses the same temporal transformer configuration as BC-DP, consisting of 4 layers, 6 attention heads, and a maximum sequence length of 10.
Action prediction is performed using a two-layer MLP head with a hidden size of 1024.
For T-MEE, the kernel bandwidth is fixed to $\sigma=0.5$, and the MEE loss weight follows the same setting as BC-RNN, with higher weights for spatial, goal, and long-horizon tasks, and a lower weight for object-centric tasks.

\subsubsection{Large-scale VLA models}
All large-scale models share the same training protocol and optimization settings unless otherwise specified.
We use separate learning rates for the vision--language backbone and the action model, and train all models with a unified optimizer, scheduler, batch size, and training budget.
The number of training steps and action chunk length differ between LIBERO and SimplerEnv.
For models equipped with an action head (except OFT), we adopt a diffusion-based action model with a DiT-B backbone.
Detailed configurations are summarized in Table~\ref{tab: shared_large_model_settings}.

\begin{table}[htbp]
\centering
\caption{Shared training and architecture settings for large-scale VLA models.}
\label{tab: shared_large_model_settings}
\vskip -0.05in
\begin{tabular}{ll}
\toprule
\textbf{Category} & \textbf{Configuration} \\
\midrule
Optimizer 
& AdamW ($\beta=(0.9,0.95)$, $\epsilon=1\times10^{-8}$, weight decay $1\times10^{-8}$) \\

Learning Rate 
& $1\times10^{-5}$ (VLM), $1\times10^{-4}$ (action expert) \\

LR Scheduler 
& Cosine with minimum LR $1\times10^{-6}$ \\

Warm-up Steps 
& 5k steps \\

Training Steps 
& 30k (LIBERO), 40k (SimplerEnv) \\

Batch Size 
& 128 (8 GPUs $\times$ 16 samples per GPU) \\

Action Chunk Length 
& 8 (LIBERO), 16 (SimplerEnv) \\

Action Head Type 
& Diffusion-based (DiT-B), except OFT \\

Diffusion Steps 
& 4 repeated steps, 1000 timestep buckets \\

DiT Backbone 
& 16 layers, 12 heads, head dim 64, input dim 768 \\

Cross-Attention Dim 
& 2048 \\

Dropout 
& 0.2 (final-layer dropout enabled) \\

Action Hidden Dim 
& 1024 \\

Evaluation & 50 rollouts per task (LIBERO), 24 rollouts per task (SimplerEnv)\\
\bottomrule
\end{tabular}
\end{table}

\textbf{GR00T} conditions the action expert on the final-layer vision-language features produced by the backbone, using these features as the sole conditioning signal for action generation.

\textbf{PI} conditions the action expert on multi-layer VL features, aggregating representations from multiple backbone layers rather than relying only on the final-layer output.

\textbf{DS-VLA} similarly conditions the action expert on the final-layer VL features.
In addition, the image encoder used in its System~1 branch is DINOv2-ViT-S/14~\cite{oquab2024dinov2}.

\textbf{OFT} augments the VLA backbone by appending learnable tokens corresponding to the action chunk length multiplied by the action dimensionality.
These tokens are jointly optimized with the backbone to directly model temporally chunked action representations.

\subsubsection{Real-world Experiments}
We adopt a single baseline, namely the original GR00T N1.5~\cite{bjorck2025gr00t} implementation with its default architecture and a continuous-action flow-matching head.

\textbf{Shared Settings.}
For real-robot experiments, we train GR00T with an action chunk size of $K=25$ and batch size 128 on a single H100 GPU.
We follow the recommended optimization hyperparameters, using AdamW with a learning rate $1\times 10^{-4}$, weight decay $1\times 10^{-5}$, and a warmup ratio of 0.05 of the total training steps.
Consistent with the original setup, we freeze both the language model backbone and the vision tower, and fine-tune only the projector and diffusion policy head.

\textbf{MEE-based Settings.}
During training, we adopt a two-phase optimization strategy with a total of 10k training steps.
In the first one-third of the training process, the model is optimized using only the MSE objective to stabilize the action error distribution.
In the remaining two-thirds of training, we activate the MEE-based objectives.
For T-MEE, the kernel bandwidth $\sigma$ is fixed to $0.5$, and the loss weight $\alpha$ is set to $1.0$.

\subsection{Additional Experiments with Original GR00T N1.5 Models}
\label{subsec: appendix_gr00t_n1.5}

We further evaluate T-MEE on the original GR00T N1.5 models~\cite{bjorck2025gr00t} without architectural or training modifications.
As shown in Table~\ref{tab: libero_gr00t_n15}, incorporating T-MEE consistently improves performance across all LIBERO task suites, leading to a higher overall average success rate.
These results demonstrate that the effectiveness of T-MEE is not limited to re-trained or modified backbones, but also extends to strong off-the-shelf VLA models.
This further confirms the general applicability of T-MEE as a lightweight and architecture-agnostic supervision objective.

\begin{table}[htbp]
\centering
\caption{Performance comparison on LIBERO with original GR00T N1.5. Results are reported as success rates.}
\label{tab: libero_gr00t_n15}
\vskip -0.05in
\begin{tabular}{lccccc}
\toprule
Method & Spatial & Goal & Object & Long & Avg \\
\midrule
GR00T N1.5~\cite{bjorck2025gr00t}
& 93.4 & 85.0 & 89.4 & 78.4 & 86.6 \\
\rowcolor{paperbg}
\quad + T-MEE
& \textbf{93.8} & \textbf{87.2} & \textbf{93.2} & \textbf{80.4} & \textbf{88.7} \\
\bottomrule
\end{tabular}
\end{table}

\subsection{Additional Experiments with Few-shot Learning Across Models}
\label{subsec: appendix_few-shot}

Table~\ref{tab: fewshot_02} presents few-shot evaluation results on LIBERO under a fixed training ratio of 0.2 across multiple VLA architectures.
Across all models, incorporating T-MEE consistently improves the overall average success rate, demonstrating that the benefits of T-MEE are not tied to a specific backbone or action head.
These results further confirm that T-MEE serves as a generally applicable, architecture-agnostic supervision objective that enhances data efficiency in few-shot learning settings.

\begin{table}[t]
\centering
\caption{Few-shot evaluation on LIBERO under a 0.2 training ratio.
Success rates (\%) are reported for each task suite and the overall average across different VLA architectures.}
\label{tab: fewshot_02}
\vskip -0.05in
\begin{tabular}{
>{\raggedright\arraybackslash}m{2.5cm}
>{\centering\arraybackslash}m{1.2cm}
>{\centering\arraybackslash}m{1.2cm}
>{\centering\arraybackslash}m{1.2cm}
>{\centering\arraybackslash}m{1.2cm}
>{\centering\arraybackslash}m{1.2cm}
}
\toprule
Method & Spatial & Goal & Object & Long & Avg \\
\midrule
BC-Transformer~\cite{liu2023libero} 
& 42.4 & 51.0 & 57.6 & 8.8 & 40.0 \\
\rowcolor{paperbg}
\quad + T-MEE 
& \textbf{48.6} & \textbf{57.4} & \textbf{67.6} & \textbf{9.4} & \textbf{45.8} \\
\midrule
GR00T~\cite{bjorck2025gr00t} 
& \textbf{85.0} & 77.2 & 94.4 & 53.0 & 77.4 \\
\rowcolor{paperbg}
\quad + T-MEE 
& 84.8 & \textbf{79.0} & \textbf{97.8} & \textbf{60.2} & \textbf{80.5} \\
\midrule
OFT~\cite{kim2025fine} 
& 93.4 & 86.4 & 92.0 & 58.4 & 82.6 \\
\rowcolor{paperbg}
\quad + T-MEE 
& \textbf{93.4} & \textbf{88.0} & \textbf{96.4} & \textbf{59.6} & \textbf{84.4} \\
\midrule
$\pi_0$~\cite{black2024pi0} 
& 90.4 & 77.2 & 93.6 & 61.6 & 80.7 \\
\rowcolor{paperbg}
\quad + T-MEE 
& \textbf{92.4} & \textbf{82.4} & \textbf{96.0} & \textbf{63.2} & \textbf{83.5} \\
\midrule
DS-VLA~\cite{starvla2025} 
& 77.4 & 74.6 & 89.0 & 54.4 & 73.9 \\
\rowcolor{paperbg}
\quad + T-MEE 
& \textbf{81.0} & \textbf{76.8} & \textbf{96.6} & \textbf{55.0} & \textbf{77.4} \\
\bottomrule
\end{tabular}
\vskip -0.15in
\end{table}

\subsection{Additional Experiments with Noise Across Models}
\label{subsec: appendix_noise}

\subsubsection{Noise Types in Our Study}
We consider four types of noise to evaluate robustness under corrupted supervision, including two image-level corruptions and two action-level noise models.

\paragraph{Image-level Noise}
We follow the experimental setup of~\cite{li2026cronusvla} for image-level noise injection.
Specifically, \textbf{motion blur} is applied using a horizontal linear kernel with size $k=5$.
For \textbf{salt-and-pepper (SAP) noise}, a fraction of $0.2$ of image pixels is corrupted, with equal probability assigned to salt and pepper noise.
\textbf{Gaussian noise} is added with a standard deviation of $10.0$.
We next describe the three types of image-level noise considered in our experiments.

\begin{figure}[htbp]
\begin{center}
\centerline{\includegraphics[width=0.7\textwidth]{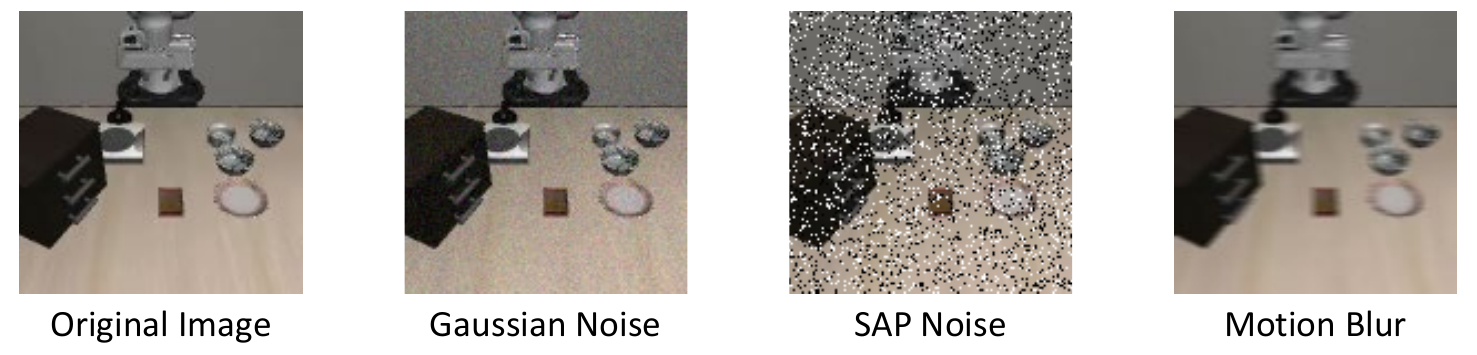}}
\caption{
Visualization of image-level noise corruptions used for robustness evaluation.
}
\label{fig: noise_types}
\end{center}
\vskip -0.15in
\end{figure}

\textbf{Motion Blur} simulates camera or object motion during image acquisition.
We apply a linear motion blur by convolving the input image with a normalized directional kernel of fixed size.
This corruption primarily degrades spatial details while preserving global structure, and represents common observation noise in real-world robotic perception.
\textbf{Salt-and-pepper (SAP) noise} randomly replaces a subset of image pixels with extreme values.
This corruption introduces sparse but high-magnitude perturbations, leading to impulsive outliers in the visual observation space.
SAP noise is commonly used to model sensor faults or transmission errors.
\textbf{Gaussian Noise} adds zero-mean Gaussian perturbations to image intensities.
Unlike SAP noise, which produces sparse and extreme outliers, Gaussian noise induces dense but small-magnitude deviations and serves as a standard baseline for modeling observation noise under Gaussian assumptions.
We visualize these three noise types in Figure~\ref{fig: noise_types}.

\paragraph{Action-level Noise}
Action-level noise perturbs the continuous action commands directly in the action space, modeling execution-time disturbances such as actuator uncertainty, control jitter, or occasional command corruption.
Formally, given a clean action $\mathbf{a} \in \mathbb{R}^d$, action-level noise produces a corrupted action $\tilde{\mathbf{a}}$ by adding stochastic perturbations to $\mathbf{a}$.
In our experiments, we consider two representative types of action-level noise: \emph{Cauchy noise}, which introduces continuous heavy-tailed perturbations, and \emph{impulse noise}, which produces sparse but extreme outliers.

\textbf{Cauchy Noise} is added to continuous action values by sampling from a Cauchy distribution.
Given a clean action $\mathbf{a} \in \mathbb{R}^d$, the corrupted action is defined as
\begin{equation}
\tilde{\mathbf{a}} = \mathbf{a} + \boldsymbol{\epsilon}, 
\quad \boldsymbol{\epsilon} \sim \mathrm{Cauchy}(\mathbf{0}, \gamma),
\end{equation}
where $\gamma$ denotes the scale parameter.
In our experiments, we set $\gamma = 0.02$.
Due to its heavy-tailed nature and undefined variance, Cauchy noise induces frequent large-magnitude perturbations, serving as a representative non-Gaussian noise model for action corruption.
The sampled noise is truncated to a bounded range in practice.

\textbf{Impulse Noise} corrupts actions by injecting large deviations with a small probability.
Specifically, the corrupted action is given by
\begin{equation}
\tilde{\mathbf{a}} =
\begin{cases}
\mathbf{a} + \boldsymbol{\delta}, & \text{with probability } p, \\
\mathbf{a}, & \text{with probability } 1 - p,
\end{cases}
\end{equation}
where $\boldsymbol{\delta}$ is sampled from a zero-mean distribution with large variance.
In our implementation, the impulse probability is set to $p = 0.05$.
Unlike Cauchy noise, which is continuously heavy-tailed, impulse noise produces sparse but extreme outliers, modeling occasional execution failures or actuator glitches.

\subsubsection{Additional Experiments}
\begin{figure}[htbp]
\begin{center}
\centerline{\includegraphics[width=0.6\textwidth]{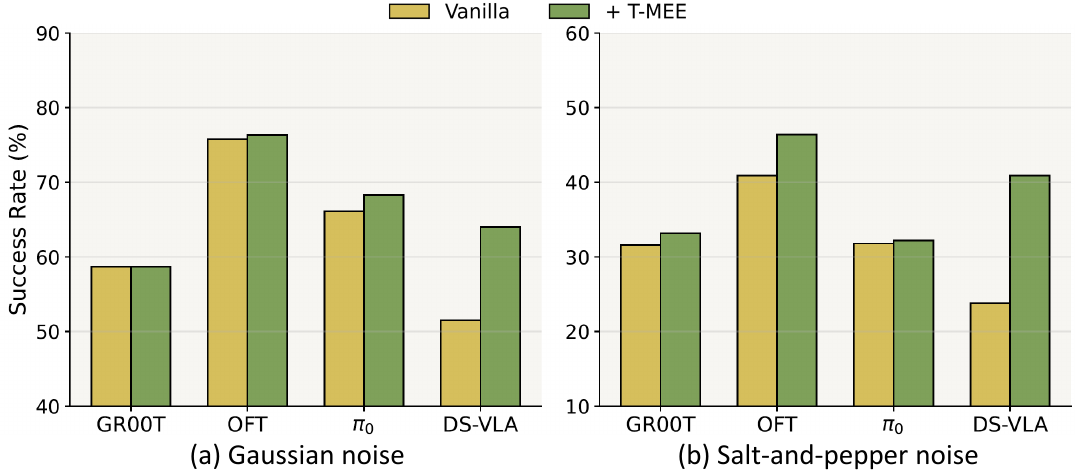}}
\caption{
Performance under image-level noise without noise-specific training on LIBERO.
Evaluation is conducted by directly testing models on noisy observations.
T-MEE improves robustness across different architectures, particularly under salt-and-pepper noise.
}
\label{fig: noise_zs}
\end{center}
\end{figure}

We further evaluate robustness by directly testing models on noisy observations without any noise-specific fine-tuning.
As shown in Figure~\ref{fig: noise_zs}, adding T-MEE consistently improves performance across different VLA architectures under both Gaussian and SAP noise.
Notably, the gains are more pronounced under SAP noise, which introduces sparse and high-magnitude perturbations.
This suggests that T-MEE enhances inherent robustness to outliers and non-Gaussian corruptions, even in the absence of explicit noise exposure during training.
Overall, these results indicate that T-MEE improves zero-shot robustness to observation noise by shaping action error distributions in a noise-agnostic manner.

\subsection{Additional Experiments with Imbalance Across Models}
\label{subsec: appendix_imbalance}

Table~\ref{tab: libero_imbalance_025} reports performance under a fixed imbalance ratio of 0.25 across multiple VLA architectures.
Across all evaluated models, incorporating T-MEE consistently improves the overall average success rate compared to the corresponding baselines.
The gains are observed across different task suites, including spatial, goal-oriented, object-centric, and long-horizon tasks, indicating that the benefits of T-MEE are not restricted to a specific architecture or task type.
These results further demonstrate that T-MEE provides a generally effective supervision signal under moderate task-level imbalance.
Combined with the main text analysis, they also suggest that the effectiveness of T-MEE holds within a practical range of imbalance, while extreme imbalance remains a challenging regime.

\begin{table}[htbp]
\centering
\caption{Performance on LIBERO under an imbalance ratio of 0.25.
Results are reported as success rates. Best results within each method block are bolded.}
\label{tab: libero_imbalance_025}
\vskip -0.05in
\begin{tabular}{
>{\raggedright\arraybackslash}m{2.5cm}
>{\centering\arraybackslash}m{1.2cm}
>{\centering\arraybackslash}m{1.2cm}
>{\centering\arraybackslash}m{1.2cm}
>{\centering\arraybackslash}m{1.2cm}
>{\centering\arraybackslash}m{1.2cm}
}
\toprule
Method & Spatial & Goal & Object & Long & Avg \\
\midrule
BC-Transformer~\cite{liu2023libero}
& 65.6 & 62.4 & 44.8 & 12.4 & 46.3 \\
\rowcolor{paperbg}
\quad + T-MEE
& \textbf{66.4} & \textbf{67.8} & \textbf{50.2} & \textbf{14.2} & \textbf{49.7} \\
\midrule
GR00T~\cite{bjorck2025gr00t}
& 95.6 & 84.6 & 98.6 & 74.6 & 88.4 \\
\rowcolor{paperbg}
\quad + T-MEE
& \textbf{95.8} & \textbf{86.2} & \textbf{99.4} & \textbf{78.2} & \textbf{89.9} \\
\midrule
OFT~\cite{kim2025fine}
& 96.6 & 86.6 & 95.8 & 79.2 & 89.6 \\
\rowcolor{paperbg}
\quad + T-MEE
& \textbf{98.0} & \textbf{89.2} & \textbf{96.0} & \textbf{82.4} & \textbf{91.4} \\
\midrule
$\pi_0$~\cite{black2024pi0}
& 97.6 & 85.2 & 99.4 & 78.4 & 90.2 \\
\rowcolor{paperbg}
\quad + T-MEE
& \textbf{97.8} & \textbf{86.2} & \textbf{99.4} & \textbf{84.2} & \textbf{91.9} \\
\midrule
DS-VLA~\cite{starvla2025}
& 89.6 & 84.2 & 99.0 & 80.4 & 88.3 \\
\rowcolor{paperbg}
\quad + T-MEE
& \textbf{96.6} & \textbf{84.8} & \textbf{99.4} & \textbf{81.4} & \textbf{90.6} \\
\bottomrule
\end{tabular}
\end{table}

\subsection{Additional Analysis Experiments}
\label{subsec: appendix_analyses}

\subsubsection{Hyperparameter Sensitivity}
\begin{figure*}[htbp]
\begin{center}
\centerline{\includegraphics[width=0.75\textwidth]{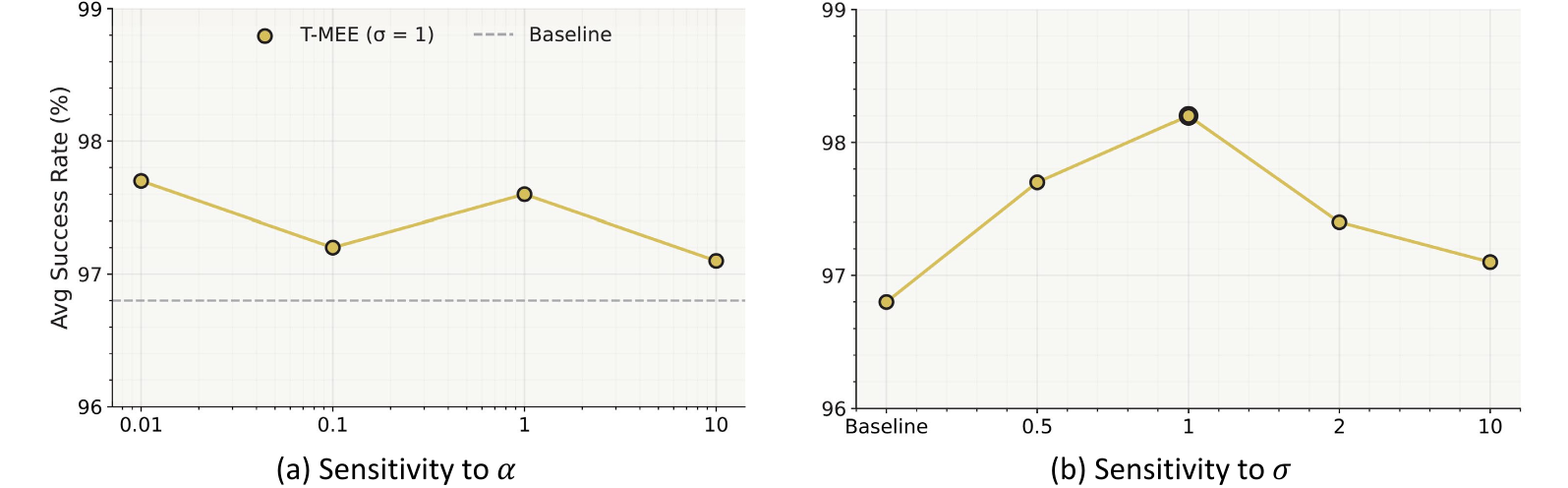}}
\vskip -0.1in
\caption{
Sensitivity analysis of T-MEE hyperparameters.
(a) Sensitivity to the loss weight $\alpha$ in Equation~\ref{eq: total_loss}.
(b) Sensitivity to the kernel bandwidth $\sigma$ in Equation~\ref{eq: t-mee}.
Results are reported as average success rates on LIBERO.
}
\label{fig: sensitivity}
\end{center}
\vskip -0.15in
\end{figure*}

Figure~\ref{fig: sensitivity} analyzes the sensitivity of T-MEE to its key hyperparameters. Overall, performance remains stable across a wide range of the loss weight $\alpha$ in Equation~\ref{eq: total_loss} and the kernel bandwidth $\sigma$ in Equation~\ref{eq: t-mee}. While extreme values lead to mild performance degradation, intermediate settings consistently yield strong results. These findings indicate that T-MEE does not require precise hyperparameter tuning and admits a broad and stable operating regime in practice.

\subsubsection{Training Overhead}
Early studies on MEE in the 2010s primarily focused on improving its computational efficiency to enable practical deployment under limited hardware resources.
In modern training regimes, however, this concern has largely diminished due to advances in GPU hardware and the substantially increased cost of large-scale model training and data loading.
In our experiments, the additional computational overhead introduced by T-MEE is negligible.
On LIBERO, we measure the per-iteration forward and backward pass time on a single NVIDIA H100 GPU to be approximately 0.59 seconds both with and without T-MEE, with no observable difference between the two settings.
In practice, overall training time is dominated by model execution, data loading, and system-level resource scheduling, rather than the entropy-based loss computation itself.
These results indicate that T-MEE incurs essentially no additional training cost in modern VLA training pipelines.

\subsection{Limitations}
While our method consistently improves performance across a range of VLA architectures and data regimes, it also introduces several limitations.
First, T-MEE introduces additional hyperparameters, including the loss weight and kernel bandwidth, as well as practical design choices such as the warm-up length before enabling the entropy objective. The need for hyperparameter tuning is a common limitation shared by many information-theoretic learning methods.
Second, trajectory-level error entropy estimation requires sufficient sample support. Under extremely imbalanced data regimes, where certain tasks are represented by only a few demonstrations, the reliability of entropy estimation degrades, and the effectiveness of T-MEE diminishes. This characterizes a practical range of imbalance within which our method remains effective, rather than providing robustness under arbitrarily severe imbalance.

\end{document}